\pdfoutput=1
\documentclass[11pt]{article}

\usepackage[preprint]{acl}

\usepackage{times}
\usepackage{latexsym}
\usepackage[T1]{fontenc}
\usepackage[utf8]{inputenc}
\usepackage{microtype}
\usepackage{inconsolata}

\usepackage{amsmath,amssymb}
\usepackage{graphicx}
\usepackage{booktabs}
\usepackage{xcolor,colortbl}
\usepackage{multirow}
\usepackage{subcaption}
\usepackage{listings}
\usepackage{enumitem}
\usepackage{bbm}
\usepackage{pifont}
\usepackage{soul}
\usepackage{xspace}
\usepackage[capitalize]{cleveref}
\usepackage{titletoc} %
\usepackage{needspace}          %

\newcommand{\cmark}{\ding{51}}
\newcommand{\xmark}{\ding{55}}
\newcommand{\expnum}[2]{{#1}\mathrm{e}{-#2}}
\newcommand{\ptok}[1]{\texttt{$\langle$#1$\rangle$}}
\definecolor{Gray}{gray}{0.9}
\lstdefinestyle{prompt}{
  basicstyle=\ttfamily\footnotesize,
  breaklines=true,
  breakatwhitespace=false,
  breakindent=0pt,
  columns=fullflexible,
  showstringspaces=false,
  upquote=true,
  xleftmargin=0.5em,
  framexleftmargin=0.5em,
  frame=leftline,
  framesep=4pt,
  rulecolor=\color{Gray},
  aboveskip=0.6em,
  belowskip=0.6em,
}
\definecolor{lblue}{rgb}{0, 0.2, 0.8}
\definecolor{orange}{rgb}{1.0, 0.5, 0.0}
\definecolor{purple}{rgb}{0.7, 0.1, 0.9}
\definecolor{yellow}{rgb}{0.75, 0.75, 0.0}
\definecolor{green}{rgb}{0, 0.8, 0.2}
\definecolor{babyblue}{rgb}{0.54, 0.81, 0.94}
\newcommand{\youngkil}[1]{#1}
\newcommand{\arredit}[1]{#1}
\definecolor{darkred}{rgb}{0.6, 0.0, 0.0}
\newcommand{\cotmark}[1]{#1}

\title{EventCoT: Event-centric Video Chain-of-thought for\\ Reasoning Temporal Localization}

\author{
  \textbf{Youngkil Song\textsuperscript{1}},
  \textbf{Yoonjae Baek\textsuperscript{1}},
  \textbf{Dongwon Kim\textsuperscript{2}},
  \textbf{Inho Kim\textsuperscript{1}},
  \textbf{Dongkeun Kim\textsuperscript{3$\dagger$}},
  \textbf{Suha Kwak\textsuperscript{1$\dagger$}}
\\
  \textsuperscript{1}Pohang University of Science and Technology, Pohang, South Korea\\
  \textsuperscript{2}Korea Advanced Institute of Science and Technology, Daejeon, South Korea\\
  \textsuperscript{3}Handong Global University, Pohang, South Korea\\
  \textsuperscript{$\dagger$}Co-corresponding authors.
}

\begin{document}
\maketitle

\begin{abstract}

\arredit{Reasoning temporal localization (RTL) requires a model to generate an answer that itself contains the time interval supporting it, coupling high-level reasoning with temporal grounding in a single response.
To tackle this challenge, we propose the first event-centric video chain-of-thought framework, dubbed EventCoT.
EventCoT first performs event-centric tokenization, converting the video into compact event tokens that enable efficient identification of question-relevant events.
It then reasons within these events to generate the answer, grounding the time interval via embedding matching that aligns placeholder tokens with visual embeddings.
EventCoT achieves state-of-the-art results on ActivityNet-RTL while using substantially fewer visual tokens than previous work, and attains strong zero-shot results on the grounded video question answering benchmark ReXTime.}
\arredit{Our code will be released for research purposes.}

\end{abstract}

\section{Introduction}
\begin{figure*}[!t]
\centering
\includegraphics[width=0.90\textwidth]{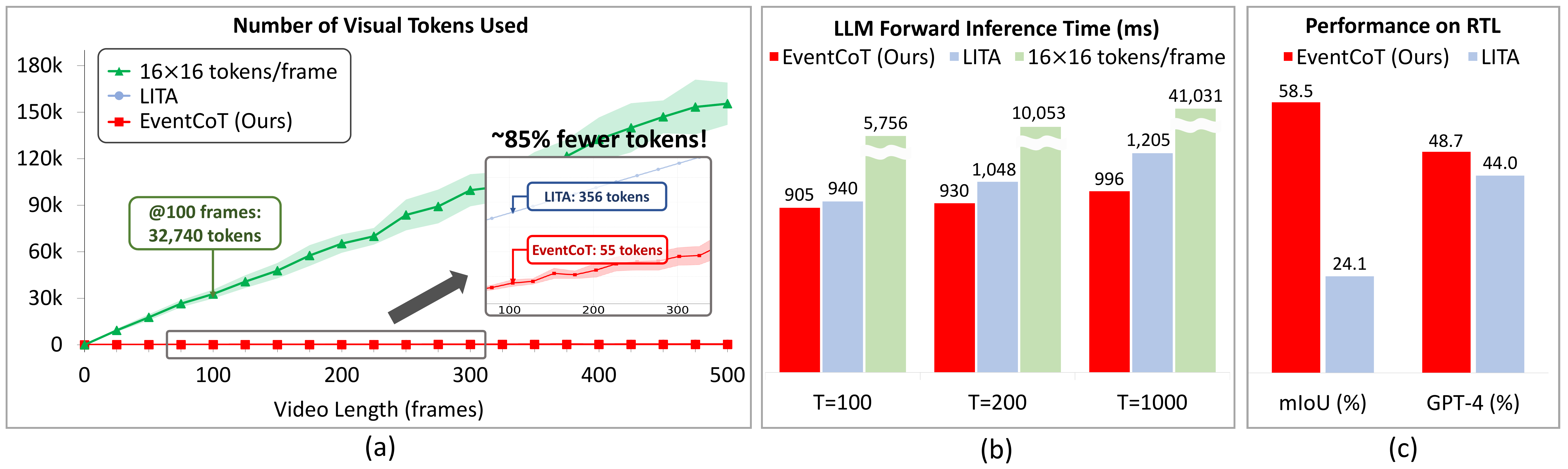}
\caption{
Advantages of using events in RTL in terms of visual token usage (a), inference time (b), and performance on ActivityNet-RTL~\citep{lita_ECCV_2024}. 
Models that process spatially detailed tokens like Temporal-CoT~\citep{temporal-cot} incur substantial LLM computation (colored in green), and query-agnostic frame sampling of LITA~\citep{lita_ECCV_2024} still introduce redundant tokens from irrelevant regions (colored in blue).
EventCoT reduces visual token usage by event-centric CoT, achieving faster inference and improved RTL performance.
}
\vspace{-4mm}
\label{fig:teaser}
\end{figure*}

\noindent
Large language models (LLMs) have demonstrated remarkable capabilities in reasoning and commonsense understanding from textual information. 
Multimodal LLMs (MLLMs) extend such capabilities to visual understanding through vision-language alignment~\citep{radford2019language, gpt4o, liu2023visual, qwen2.5-vl, llava-onevision, internvl3, mplug-owl3, vila}. 
More recently, video-LLMs~\citep{video-llava, vid2seq_CVPR_2023, qian2024momentor, videochat_arxiv_2023, zhang2023video, timechat_CVPR_2024} have been developed to reason over long-form videos and perform high-level reasoning tasks such as video question answering~\citep{wang2025lvbench, mvbench_cvpr_2024, mmbench_nips2024, hourvideo_nips2024, mlvu_cvpr2025}. 
However, most existing video-LLMs lack the precise temporal reasoning required to determine when and how long specific events occur. This limitation motivated the introduction of reasoning temporal localization (RTL)~\citep{lita_ECCV_2024}.
\youngkil{Unlike conventional video QA, where temporal grounding is at most an auxiliary output, RTL requires a model to generate an answer that itself contains the supporting time interval.
Reasoning and temporal localization therefore cannot be solved in isolation and must be produced jointly within a single response, which remains challenging even for recent large-scale foundation models.}

The pioneering work on RTL~\citep{lita_ECCV_2024} extends video-LLMs to jointly predict a time interval and an answer.
Particularly for temporal grounding, it uniformly samples frames across the entire video to ensure global temporal coverage, and exploits their features as 
input; time intervals are then estimated via the next-token prediction regime of LLMs with discrete time tokens incorporated into the LLM vocabulary.
While effective, this design suffers from two fundamental limitations.
\arredit{First, it feeds the LLM visual tokens from uniformly sampled frames regardless of the question. This query-agnostic visual tokenization introduces substantial noise when the sampling is overly dense and information loss when it is too sparse.}
\arredit{Second, time interval estimation via next-token prediction entangles temporal grounding with answer generation, which degrades both precision of temporal grounding and answer quality.}
Recent agentic systems~\citep{liu2025videomind, menon2025caviar} 
resolve the first issue by iteratively retrieving
query-relevant video segments using external tools.
While this design allows the model to condition its reasoning on selected temporal regions, it requires multiple rounds of LLM-driven planning and repeated model calls that operate on spatially detailed visual features, substantially increasing inference cost and token usage.
\arredit{Moreover, it does not address the second issue as it couples temporal grounding and answer generation within an iterative generation loop.}

\cotmark{To address these limitations, we propose EventCoT, an event-centric video framework for RTL. Chain-of-thought (CoT) reasoning~\citep{wei2022chain} solves a complex problem by generating intermediate reasoning steps before the final answer, rather than producing the answer directly. EventCoT follows this principle. It first selects the question-relevant events as an intermediate reasoning step, and then reasons within these events to produce the temporally grounded answer. We realize this chain of thought as a two-step operation.}
\youngkil{\arredit{In Step~1, EventCoT selects the question-relevant events using compact event tokens produced by an \emph{event tokenizer}.}
\arredit{We take inspiration from human perception, where continuous activity is understood as a sequence of discrete events~\citep{zacks2007event}, and the event tokenizer segments the video into non-overlapping events and summarizes each as a compact representation.}
This removes the redundancy of query-agnostic uniform tokenization and lets the model reason over events rather than individual frames\arredit{, as demonstrated in Fig.~\ref{fig:teaser}}.
\arredit{In Step~2, EventCoT performs the fine-grained reasoning that grounds the temporal interval and generates the answer, relying on \emph{embedding matching}.}
\arredit{Instead of predicting the interval with discrete time tokens, which entangles temporal grounding with answer generation, EventCoT aligns the placeholder tokens \ptok{segment\_key}, \ptok{start}, and \ptok{end} with visual features by semantic similarity.}
\arredit{Because this matching operates in both steps, temporal grounding stays disentangled from answer generation, so that each sub-task is optimized without competing with the other.}}

\youngkil{We evaluate EventCoT on ActivityNet-RTL~\citep{lita_ECCV_2024} for reasoning temporal localization and on ReXTime~\citep{chen2024rextime} for grounded video question answering.}
\youngkil{EventCoT achieves state-of-the-art results on ActivityNet-RTL and strong zero-shot performance on ReXTime, while using substantially fewer visual tokens.}
\arredit{These results demonstrate that event-centric chain-of-thought reasoning improves both temporal grounding and answering performance, and also reduces the inference time of the overall reasoning process.}
In summary, our contribution is three-fold:
\begin{itemize}[leftmargin=5mm] 
    \setlength\itemsep{1mm}
    \item \arredit{We propose EventCoT, the first event-centric chain-of-thought framework that decouples temporal grounding and answer generation for efficient reasoning temporal localization (RTL).}
    \item To reduce visual token redundancy while enabling temporally grounded reasoning, we introduce an event tokenizer and an embedding matching mechanism, and verify their effectiveness through extensive experiments.
    \item \youngkil{EventCoT achieves state-of-the-art results on ActivityNet-RTL and strong zero-shot performance on ReXTime, while using substantially fewer visual tokens.}
\end{itemize}

\section{Related Work}
\begin{figure*}[t]
    \centering
    \includegraphics[width=0.91\textwidth]{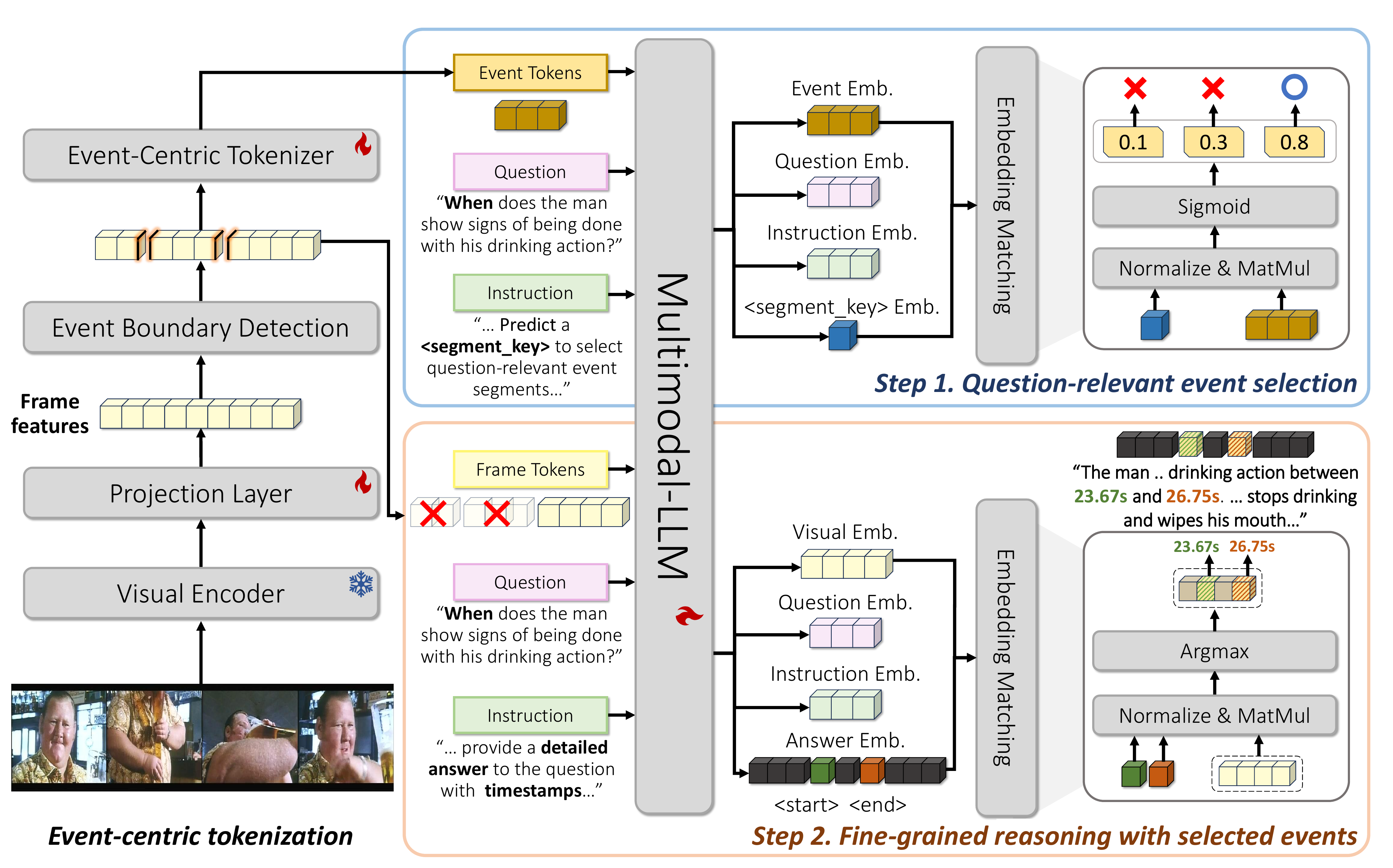}
    \caption{Overview of EventCoT.
    EventCoT first tokenizes a video into compact event tokens via event-centric tokenization.
    Given the question, it selects question-relevant events by matching the \texttt{$\langle$segment\_key$\rangle$} token to event tokens (Step~1).
    It then performs fine-grained reasoning on frames from the selected events and predicts \texttt{$\langle$start$\rangle$} and \texttt{$\langle$end$\rangle$} to generate temporally grounded answers (Step~2).}
    \label{fig:main_fig}
    \vspace{-4mm}
\end{figure*}

\noindent \youngkil{\textbf{Grounded Video Question Answering.}
Benchmarks such as TVQA+~\citep{tvqaplus}, NExT-GQA~\citep{nextgqa}, and ReXTime~\citep{chen2024rextime} evaluate whether an answer is supported by temporal evidence, asking a model to predict the supporting time span separately from its multiple-choice answer.
\arredit{In these protocols, the answer itself carries no temporal extent, and correctness and localization are combined post hoc through metrics such as Acc@GQA, so a model need not embed temporal evidence in its generated answer.}
\arredit{Reasoning temporal localization~\citep{lita_ECCV_2024} instead defines the answer as the start and end timestamps with the explanation, so reasoning and temporal localization must be produced jointly in a single response.}
\arredit{This joint generation makes RTL more challenging than grounded VQA and motivates our design disentangling the two through embedding matching.}}

\noindent \textbf{Event-centric Video Understanding.}
\youngkil{\arredit{TRACE \citep{guo2024trace} structures the LLM's output as a sequence of event triplets to cast temporal grounding as autoregressive generation,} and Chat-UniVi~\citep{jin2024chat} merges visual tokens within segmented events to compress visual redundancy.
However, both restrict the role of events to output formatting or passive token compression, without selecting or reasoning over question-relevant events.
In contrast, EventCoT uses events as the unit of both input representation and temporal reasoning, isolating question-relevant events first and performing fine-grained temporal grounding within them.}

\noindent \textbf{Chain-of-Thought Reasoning in VLMs. }
\youngkil{Chain-of-thought (CoT) reasoning~\citep{wei2022chain} decomposes a complex problem into intermediate reasoning steps that precede the final answer.
In VLMs, these intermediate steps need not be textual.
Visual-CoT~\citep{visual-cot} first predicts bounding boxes of key regions as intermediate cues, Temporal-CoT~\citep{temporal-cot} iteratively selects the most relevant frames of a long video before final reasoning, and agentic systems~\citep{wang2024videoagent, zhi2025videoagent2} realize similar iterations by orchestrating external tools.
\arredit{These works establish that selecting visual units can itself serve as an intermediate reasoning step, yet the iterative frame selection mechanism remains computationally inefficient and suffers from token redundancy.}
\arredit{EventCoT follows the same principle with events as the unit of intermediate reasoning, first selecting question-relevant events at a coarse level (Sec.~\ref{sec:step1}) and then performing fine-grained reasoning within them (Sec.~\ref{sec:step2}), preserving the CoT structure without iterative frame selection.}}

\noindent \textbf{VLMs for Grounded and Specialized Outputs.}
\arredit{While most VLMs generate only text, recent approaches enable grounded, non-textual outputs by mapping the hidden embeddings of dedicated placeholder tokens to task-specific heads, keeping high-level reasoning in the LLM.}
\arredit{LISA~\citep{lai2024lisa} passes the embedding of a special \texttt{<SEG>} token to a mask decoder to produce segmentation masks for implicit queries, and ETBench~\citep{liu2024bench} matches the embedding of a \texttt{<vid>} token against frame features to predict precise timestamps.}
\arredit{Building on this idea, we develop two-step embedding matching using placeholder embeddings for event selection and temporal grounding.}

\section{Proposed Method}
\label{sec:method}

This section introduces EventCoT, the first event-centric video chain-of-thought (CoT) framework for reasoning temporal localization (RTL).
As shown in Fig.~\ref{fig:main_fig}, 
EventCoT first represents input video as a compact set of event token embeddings (Sec.~\ref{sec:step0}), and then performs its \cotmark{CoT process} for RTL: \arredit{identifying events relevant to the input question (Sec.~\ref{sec:step1}), followed by fine-grained reasoning within the identified events to generate answers with temporal localization (Sec.~\ref{sec:step2}).}

\subsection{Event-centric Tokenization}
\label{sec:step0}
For event-centric tokenization of an input video, EventCoT first detects event boundaries from frame-wise features and then extracts features of the events defined by the detected boundaries. 

\subsubsection{Frame Feature Extraction. }

Given an input video, we uniformly sample $T$ frames and extract frame-wise features using a frozen CLIP-L/14~\citep{clip_pmlr21} visual encoder. 
A projection layer followed by spatial mean pooling yields 
$\mathbf{F} = [\mathbf{f}_1, \ldots, \mathbf{f}_T]^\top \in \mathbb{R}^{T \times D}$,
where $\mathbf{f}_i \in \mathbb{R}^{D}$ denotes the frame-wise feature of the $i$-th frame.
Finally, we apply rotary positional embeddings~\citep{su2024roformer} to the frame features, allowing the model to capture relative temporal dependencies across time.

\subsubsection{Event Boundary Detection. }
\label{sec:event_boundary_detection}
\arredit{To efficiently represent videos,
we
divide the frame sequence into
events, where
ideally an event means a time interval that cannot be further split in terms of semantics.}
The key is to accurately detect event boundaries,
where semantic or visual changes occur.
\arredit{A na\"ive approach is to treat frames with a large average distance to their $k$-nearest neighbors as boundary candidates.
However, such distances alone are unreliable, as frames in highly dynamic scenes appear distinct even within the same event, causing false alarms while missing subtle transitions in static scenes.}

\begin{figure}[t]
\centering
\includegraphics[width=\columnwidth]{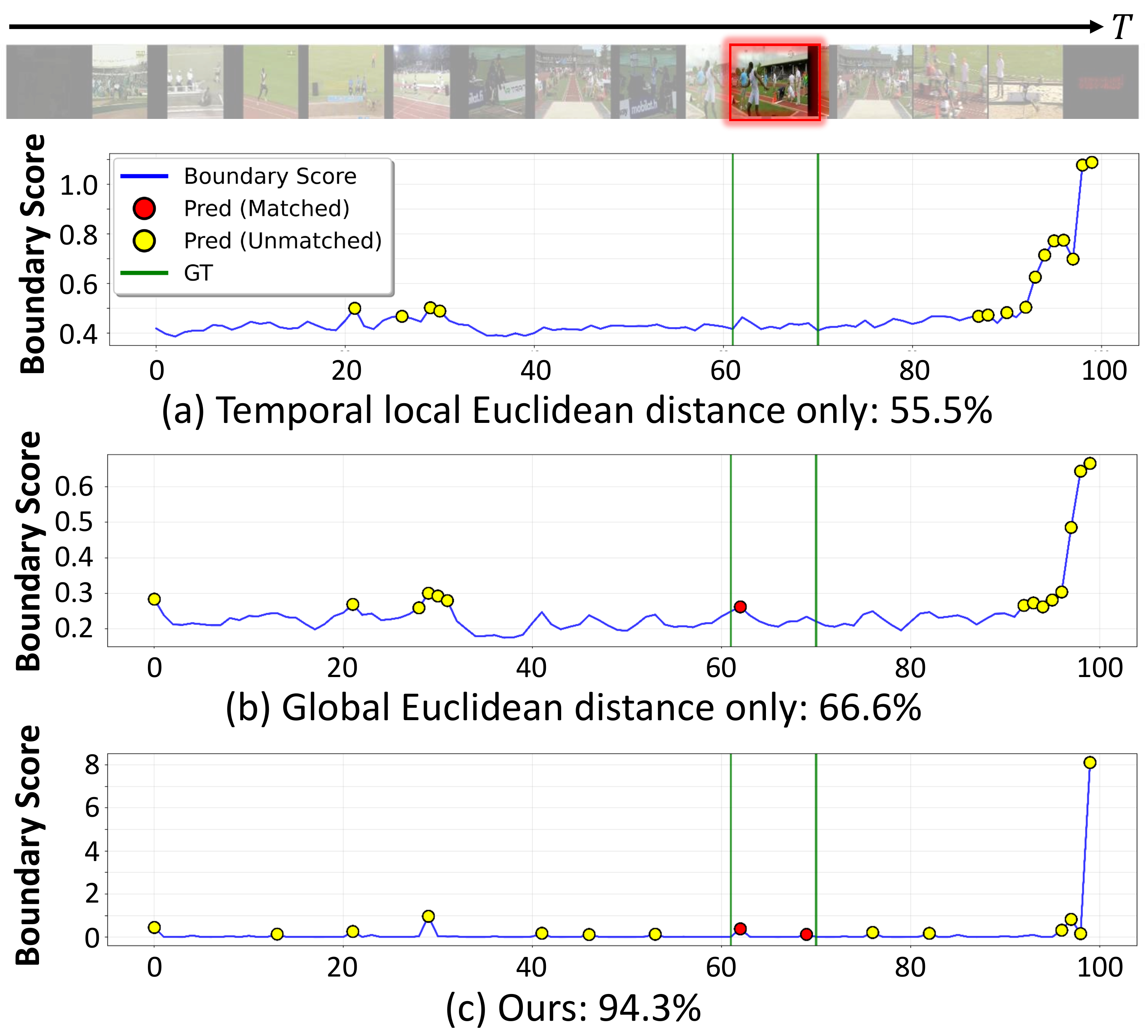}
\vspace{-6mm}
\caption{
Event boundary detection results on ActivityNet-RTL~\citep{lita_ECCV_2024}.
Numerical values indicate boundary recall computed within a tolerance of $\pm5\%$ of the video duration.}
\vspace{-4mm}
\label{fig:boundary_comparison}
\end{figure}

To resolve this issue, we adopt the idea of density peak clustering with $k$-nearest neighbors (DPC-KNN)~\citep{dpc-knn}, which suppresses unstable peaks by considering both relative distance and local density. Specifically, we introduce two key improvements tailored for event
boundary detection: temporal proximity-aware distance and power-scaled distance factor.
When computing pairwise distances between frames, we consider feature distance and temporal proximity at once:
\begin{equation}
\label{eq:combined_distance}
d(i, j) = (1 - \alpha) \|\mathbf{f}_i - \mathbf{f}_j\|_2 + \alpha \frac{|i - j|}{T},
\end{equation}
where $\alpha=0.3$ controls the temporal proximity weight and $T$ is the total number of frames.
Then, the local boundary score is estimated as the mean distance to its $k$-nearest neighbors:
\begin{equation}
\label{eq:local_boundary}
\rho_i = \frac{1}{k} \sum_{j \in \mathcal{N}_k(i)} d(i, j),
\end{equation}
where $\mathcal{N}_k(i)$ is the set of the $k$-nearest neighbors based on $d(i, j)$;
higher $\rho_i$ indicates potential boundaries.
Next, the power-scaled distance factor is applied to further suppress noisy peaks.
\arredit{The original distance factor $\delta$ quantifies the relative isolation of each frame as the distance to its nearest frame with a higher local score $\rho$, assigning small values to non-peak frames and large values only to local density maxima, yielding fewer and more meaningful peaks per cluster.
Its formal definition is provided in Appendix~\ref{supple:method_details}.}

To further strengthen this effect, we apply normalization followed by power scaling:
\begin{equation}
\label{eq:power_scaling}
\tilde{\delta}_i = \left[\text{Normalize}(\delta_i)\right]^p,
\end{equation}
\arredit{where the normalization uses the 10th and 90th percentiles for robustness against outliers, and $p=2$.}
\arredit{This scaling amplifies large $\delta$ values while suppressing small ones, enforcing sharper peaks and yielding more stable boundary detection.}

The final boundary score is defined as 
$b_i = \rho_i \cdot \tilde{\delta}_i$.
We then select the top $N-1$ frames with the highest scores as boundaries, partitioning the video into $N$ events $\{\mathcal{E}_n\}_{n=1}^{N}$.
\arredit{As shown in Fig.~\ref{fig:boundary_comparison}, our technique detects boundaries in both dynamic and static scenes, whereas those using temporally adjacent frame distances or nearest-neighbor feature distances fail to do so.}
\arredit{Qualitative examples of event segmentation are provided in Appendix~\ref{supple:qual}.}

\subsubsection{Per-event Embedding. }
\label{sec:per-event embedding}
\begin{figure}[t]
\centering
\includegraphics[width=\columnwidth]{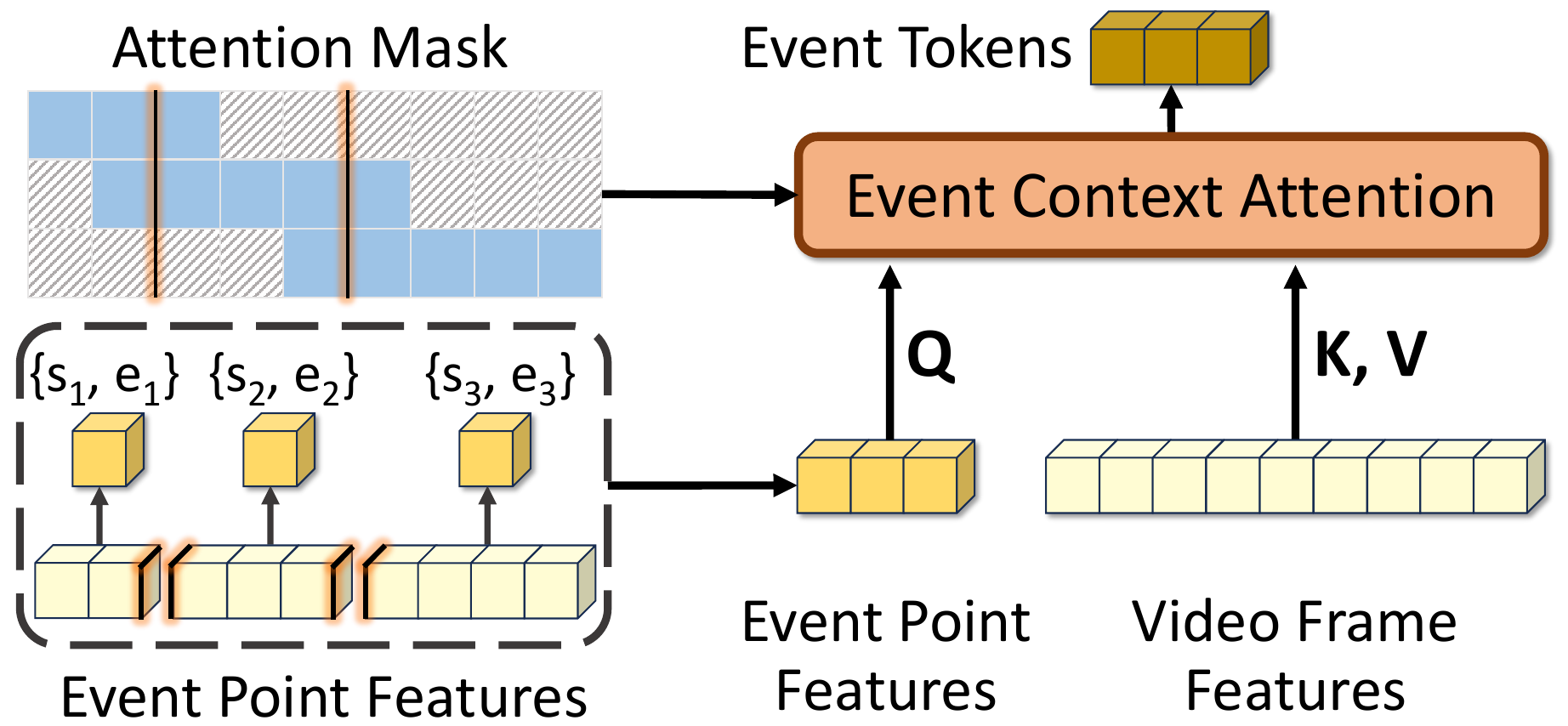}
\vspace{-6mm}
\caption{Details of the event-centric tokenizer.
By using an attention mask over an extended event boundary region, each event point feature aggregates local frames to form an event token.}
\vspace{-5mm}
\label{figs:event_centric_tokenizer}
\end{figure}

To obtain compact and discriminative representations for each event, we propose an event-centric tokenizer (Fig.~\ref{figs:event_centric_tokenizer}), inspired by point-to-region attention~\citep{li2025llava}.
\arredit{For each event $\mathcal{E}_n$, we first compute an event point feature $\mathbf{p}_n = \frac{1}{|\mathcal{E}_n|} \sum_{i \in \mathcal{E}_n} \mathbf{f}_i$ by mean pooling its frame-wise features, which provides a coarse summary of the event.}
We then refine it using \emph{event context attention}: $\mathbf{p}_n$ serves as a query, while frame-wise features $\textbf{F}$ act as keys and values, producing an event token that captures the context
within its interval.
To mitigate potential boundary inaccuracies, we expand the attention scope by including $\beta$ frames before and after each interval:
\begin{equation}
\label{eq:event_feature}
\mathbf{e}_n = \text{Attention}(\mathbf{p}_n, \{\mathbf{f}_i \mid i \in \mathcal{E}_n^{\text{ext}}\}),
\end{equation}
\arredit{where $\mathcal{E}_n^{\text{ext}}$ denotes the frame indices of the $n$-th event extended by $\beta$ frames on each side within the video range, as defined in Appendix~\ref{supple:method_details}.}
The resulting event tokens $\{\mathbf{e}_n\}_{n=1}^{N}$ are used 
as input to the LLM for question-relevant event selection.

\subsection{Step 1: Question-relevant Event Selection}
\label{sec:step1}

The first step identifies question-relevant events. 
To this end, we introduce a placeholder token \ptok{segment\_key} into the LLM vocabulary,
which serves to encode both the question and the visual context.
When this token is processed, the LLM produces the final-layer hidden embedding $\mathbf{h}_{\text{sk}}$ at its position. 
This embedding acts as a query that allows the model to determine which events should be selected.
For each event token, we also extract its final-layer hidden embedding $\mathbf{h}_{\mathbf{e}_n}$ and compute its relevance score $r_n$ via embedding matching:
\begin{equation}
\label{eq:segment_selection}
r_n = \sigma\left(\frac{\mathbf{h}_{\text{sk}}^{\phantom{0}\top} \mathbf{h}_{\mathbf{e}_n}}{\tau}\right),
\end{equation}
where $\tau$ is the temperature, and $\sigma$ is the sigmoid function. 
Finally, events with $r_n > \gamma \cdot \max_n r_n$ are selected as the question-relevant ones.

\subsection{Step 2: Fine-grained Reasoning with Relevant Events}
\label{sec:step2}

The second step performs 
reasoning conditioned on the selected question-relevant events to predict accurate timestamps and generate textual answers.

\subsubsection{Frame-level Visual Token Generation.}
We first construct frame-level visual tokens from the relevant events.
During training, the events that overlap with the ground-truth interval are directly provided as the relevant ones, rather than relying on the prediction from Step~1.
This isolates Step~2 training from potential event selection errors in Step~1.
\arredit{However, using only the overlapping events creates a shortcut, since the selection implicitly leaks ground-truth timestamps; the model can overfit to simply predict interval boundaries as the start and end timestamps, rather than performing genuine temporal reasoning.}
\arredit{To prevent this, we augment the visual tokens with a randomly sampled subset of unselected events, which breaks the boundary alignment between the input intervals and the ground-truth interval, encouraging robust temporal grounding.}
Formally, visual tokens input to the LLM for Step 2 are denoted as:
\begin{equation}
\label{eq:visual_tokens}
\mathbf{V} = \{\mathbf{f}_i \mid i \in \mathcal{E}_n, n \in \mathcal{S} \cup \mathcal{R}\},
\end{equation}
\arredit{where $\mathcal{S}$ denotes the selected event indices, and $\mathcal{R}$ is a random subset of unselected events.}
\youngkil{We also apply this random sampling at inference so that the input matches the training distribution. The full inference procedure is detailed in Appendix~\ref{supple:process_detail}.}

\subsubsection{Answer Generation with Temporal Grounding.}

In this stage, EventCoT generates an answer conditioned on the question and the constructed visual tokens $\mathbf{V}$.
Specifically, we introduce two placeholder tokens, \ptok{start} and \ptok{end}, into the LLM vocabulary to represent the temporal boundaries 
of the interval supporting
the answer.
The LLM auto-regressively generates the textual descriptions. 
At each placeholder token's position, we extract the hidden embedding and perform embedding matching with frame-wise features to predict the corresponding timestamp~\citep{liu2024bench}.
For the start time prediction, we compute the similarity between the LLM's final-layer hidden embedding for \ptok{start}, denoted by $\mathbf{h}_{\text{ts}}$, and that for each frame-wise feature $\mathbf{f}_i$ in $\mathbf{V}$ of Eq.~\eqref{eq:visual_tokens}, denoted by $\mathbf{h}_{\mathbf{f}_i}$:
\begin{equation}
\label{eq:timestamp_matching}
p_i = \text{softmax}\left(\frac{\mathbf{h}_{\text{ts}}^{\phantom{0}\top}  \mathbf{h}_{\mathbf{f}_i}}{\tau}\right).
\end{equation}
The frame with the highest similarity score is selected, converted to a timestamp, and substituted into the placeholder to form the final answer.
The end time prediction follows the same process using the embedding of \ptok{end}, $\mathbf{h}_\text{te}$.

\subsection{Training Objectives}
\label{sec:training}

\subsubsection{Event Selection Loss. }

In the first step, we train the model with the standard language modeling loss $\mathcal{L}_{\text{lm}}$ and the event selection loss $\mathcal{L}_{\text{sel}}$. 
For each event, we assign a binary label based on its temporal overlap with the ground-truth interval: events that overlap with a ground-truth interval by at least one frame are labeled as positive, and all others as negative. 
The event selection loss $\mathcal{L}_{\text{sel}}$ is defined as the binary cross-entropy (BCE) loss between the predicted relevance score $r_n$ and the binary label $y_n \in \{0, 1\}$.

\subsubsection{Temporal Localization Loss. }
In the second step, 
our model learns 
timestamp prediction using both a cross-entropy (CE) loss with Gaussian-smoothed labels and a DIoU loss for direct interval regression. 
For each frame $i$, a Gaussian-smoothed label is computed by
\begin{equation}
\label{eq:gaussian_label}
g_i = \exp\left(-\frac{(t_i - t_{\text{gt}})^2}{2\sigma^2}\right),
\end{equation}
where $t_i$ is the timestamp of frame $i$, $t_{\text{gt}}$ is the ground truth timestamp, and $\sigma=0.05$. 
The label is normalized via softmax as $\tilde{y}_i = g_i / \sum_j g_j$. 
We then compute the cross-entropy loss $\mathcal{L}_{\text{ce}}$ between the predicted probability $p_i$ (Eq.~\eqref{eq:timestamp_matching}) and the normalized target distribution $\tilde{y}_i$.
Additionally, we compute the predicted interval as the weighted sum of frame timestamps:
\begin{equation}
\label{eq:pred_interval}
\hat{t}^{\text{start}} = \sum_i p_i^{\text{start}} \cdot t_i, \quad \hat{t}^{\text{end}} = \sum_i p_i^{\text{end}} \cdot t_i,
\end{equation}
\arredit{and apply DIoU loss $\mathcal{L}_{\text{diou}}$~\citep{zheng2020distance} between the predicted interval $[\hat{t}^{\text{start}}, \hat{t}^{\text{end}}]$ and the ground-truth interval, following ActionFormer~\citep{zhang2022actionformer} and MATR~\citep{song2024online}, whose formal definition is provided in Appendix~\ref{supple:method_details}.
The timestamp matching loss combines both terms as $\mathcal{L}_{\text{ts}} = \mathcal{L}_{\text{ce}} + \mathcal{L}_{\text{diou}}$.}

EventCoT is trained end-to-end in a multi-turn setting, where event selection and fine-grained reasoning are optimized jointly.
The overall training objective is:
$\mathcal{L} = \mathcal{L}_{\text{lm}} + \lambda_{\text{sel}} \cdot \mathcal{L}_{\text{sel}} + \lambda_{\text{ts}} \cdot \mathcal{L}_{\text{ts}}$,
where $\lambda_{\text{sel}}$ and $\lambda_{\text{ts}}$ denote loss weights.
\section{Experiments}
\label{sec:experiments}

\subsection{Experimental Settings}
\noindent
\textbf{Training Datasets.}
\arredit{EventCoT is trained on a mixture of video-language datasets with and without temporal annotations following the training setup of LITA~\citep{lita_ECCV_2024}.
The former include ActivityNet-RTL~\citep{lita_ECCV_2024}, ActivityNet-Captions~\citep{krishna2017dense}, and YouCook2~\citep{zhou2018towards}, while the latter include NExT-QA~\citep{xiao2021next} and LLaVA-150K~\citep{liu2023visual}.
Dataset details and training formulations are provided in Appendix~\ref{supple:exp_details}.}

\begin{table*}[t]
\centering
\caption{
Quantitative comparisons in RTL performance
on ActivityNet-RTL~\cite{lita_ECCV_2024}.
Visual token usage of agentic methods
is not fixed per query, and thus
we report the maximum role-wise token budget for VideoMind and the approximate number of frames per tool call for CAViAR.}
\vspace{-2mm}
\label{tab:lita_benchmark}
\resizebox{0.98\textwidth}{!}{
\begin{tabular}{lcccccccc}
\toprule
\textbf{Method} & \textbf{Backbone} & \textbf{Fine-tuning} & \textbf{\# Visual Tokens} & \textbf{mIoU} & \textbf{P@0.3} & \textbf{P@0.5} & \textbf{P@0.7} & \textbf{GPT-4} \\
\midrule
Qwen3.5-9B~\cite{qwen3.5} & Qwen3.5-9B & \xmark & 7{,}150 & 37.1 & 47.6 & 37.6 & 24.0 & 39.9 \\
TimeLens-8B~\cite{timelens} & Qwen3-VL-8B & \xmark & 4{,}000 & 34.7 & 49.8 & 31.9 & 21.0 & 41.9 \\
Gemini 2.5 Pro~\cite{gemini2.5} & - & \xmark & $\sim$26.7K & 30.2 & 40.6 & 27.5 & 17.9 & 30.6 \\
Gemini 2.5 Flash~\cite{gemini2.5} & - & \xmark & $\sim$25.9K & 28.6 & 40.2 & 24.9 & 15.7 & 41.1 \\
GPT-5~\cite{gpt5} & - & \xmark & 7{,}104 & 23.7 & 35.4 & 15.7 & 7.0 & 32.3 \\
GPT-4o~\cite{gpt4o} & - & \xmark & 8{,}605 & 20.2 & 29.3 & 13.1 & 7.0 & 29.9 \\
GPT-5-mini~\cite{gpt5} & - & \xmark & 9{,}804 & 19.9 & 28.4 & 15.3 & 6.6 & 28.1 \\
\midrule
Video-LLaMA-v2~\cite{videollama2} & Vicuna-13B & \cmark & 32 & - & - & - & - & 32.1 \\
Video-ChatGPT~\cite{maaz2024video} & Vicuna-13B & \cmark & 356 & - & - & - & - & 38.8 \\
LITA-7B~\cite{lita_ECCV_2024} & Vicuna-7B & \cmark & 356 & 24.1 & 35.6 & 21.2 & 9.2 & 44.0 \\
LITA-13B~\cite{lita_ECCV_2024} & Vicuna-13B & \cmark & 356 & 28.8 & - & 25.6 & - & \underline{46.3} \\
VideoMind-7B~\cite{liu2025videomind} & Qwen2-VL-7B & \cmark & up to $9.6\text{K}$ / role & 31.3 & - & 28.0 & - & - \\
CAViAR~\cite{menon2025caviar} & Gemini-1.5 Flash & \cmark & $\approx 120$ frames / call & 32.3 & - & - & - & - \\
Qwen3.5-9B~\cite{qwen3.5} & Qwen3.5-9B & \cmark & 7{,}150 & 42.1 & 60.7 & 43.2 & 25.3 & 38.0 \\
TimeLens-8B~\cite{timelens} & Qwen3-VL-8B & \cmark & 4{,}000 & 40.1 & 58.1 & 41.9 & 25.8 & 34.6 \\
\midrule
\multicolumn{9}{l}{EventCoT (Ours)} \\
\quad Step 1: Event Selection & Vicuna-7B & \cmark & 16 & 56.2 & 76.9 & 62.0 & 45.4 & - \\
\quad Step 2: Answer Generation & Vicuna-7B & \cmark & 8--100 (39 on average) & \underline{58.5} & \underline{78.6} & \underline{66.8} & \underline{45.9} & \underline{48.7} \\
\midrule
EventCoT (Ours) & Qwen2.5-7B & \cmark & 20--100 (47 on average) & \textbf{61.4} & \textbf{84.7} & \textbf{72.1} & \textbf{46.3} & \textbf{56.9} \\
\bottomrule
\end{tabular}}
\end{table*}

\noindent
\textbf{Implementation Details.}
We use Vicuna-7B~\citep{vicuna2023} as the LLM backbone, and initialize both the LLM and the projection layer with the pretrained LLaVA weights~\citep{liu2023visual}. \arredit{To demonstrate that our method is effective across different backbones, we additionally instantiate EventCoT with a Qwen2.5-7B backbone.}
We uniformly sample $T=100$ frames per video and set the number of events to $N=16$.
\arredit{All remaining hyperparameters, listed in Appendix~\ref{supple:exp_details}, are fixed to a single configuration per backbone across every dataset and benchmark, without per-dataset tuning.}

\subsection{Evaluation Benchmarks}

\youngkil{To evaluate EventCoT, we use two main benchmarks: ActivityNet-RTL~\citep{lita_ECCV_2024} for reasoning temporal localization and ReXTime~\citep{chen2024rextime} for grounded video question answering.}
\arredit{Construction details of both benchmarks are provided in Appendix~\ref{supple:exp_details}.}
\textbf{ActivityNet-RTL}
is an open-ended benchmark that requires generating a free-form answer and the corresponding start and end timestamps for a given question and video. 
\arredit{It contains 229 curated question--answer pairs spanning 160 videos.}
\textbf{ReXTime}
\youngkil{is a grounded video question answering benchmark formulated as a multiple-choice task with four answer options, where the model selects the correct answer and separately localizes the relevant moment.} 
\arredit{It contains 921 validation samples and 2,143 test samples.}

\begin{table*}[t]
    \centering
    \caption{Zero-shot RTL performance on ReXTime~\cite{chen2024rextime}.
    Localization quality is measured by R@0.3, R@0.5, and mIoU metrics.
    Accuracy and Accuracy@IoU$\geq$0.5 indicate question answering performance.
    Best results are in \textbf{bold} and second-best are \underline{underlined}.
    }
    \vspace{-2mm}
    \label{tab:zs_rextime}
    \resizebox{0.98\textwidth}{!}{%
    \begin{tabular}{lccccccc}
        \toprule
        \textbf{Model} & \textbf{Backbone} & \textbf{\# Visual Tokens} & \textbf{R@0.3} & \textbf{R@0.5} & \textbf{mIoU} & \textbf{Acc} & \textbf{Acc@IoU} \\
        \midrule
        VTimeLLM~\cite{huang2024vtimellm} & Vicuna-7B  & 100 & 28.84 & 17.41 & 20.14 & 36.16 & -- \\
        TimeChat~\cite{timechat_CVPR_2024} & LLaMA2-7B & 3,072 (96) & 14.42 & 7.61 & 11.65 & 40.04 & -- \\
        LITA-13B~\cite{lita_ECCV_2024} & Vicuna-13B & 356 & 29.49 & 16.29 & 21.49 & 34.44 & -- \\
        TOGA~\cite{gupta2025toga} & Mistral-7B & 2,028 & 29.91 & 19.79 & 25.53 & -- & -- \\
        Qwen2.5-VL~\cite{qwen2.5-vl} & Qwen2.5-VL-7B & $\leq$24,576 & 16.05 & 9.24 & 13.60 & 56.60 & 6.35 \\
        GraphThinker~\cite{cheng2026graphthinker} & Qwen2.5-VL-7B & -- & 33.92 & 20.25 & 25.34 & \underline{66.82} & 15.21 \\
        \midrule
        EventCoT (Ours) & Vicuna-7B & 51--100 (81 on average) & \underline{46.80} & \underline{33.33} & \underline{36.50} & 48.21 & \underline{18.1} \\
        EventCoT (Ours) & Qwen2.5-7B & 1--100 (72 on average) & \textbf{51.90} & \textbf{37.57} & \textbf{39.90} & \textbf{68.5} & \textbf{27.3} \\
        \bottomrule
    \end{tabular}}
    \vspace{-3mm}
\end{table*}

\begin{table*}[t]
\centering

\begin{minipage}[t]{0.49\linewidth}
\centering
\caption{
Comparisons between the uniformly divided intervals and our events in terms of temporal localization quality.
}
\vspace{-2mm}
\label{tab:uniform_vs_event}
\resizebox{\linewidth}{!}{
\begin{tabular}{lccccc}
\toprule
\textbf{Method} & \textbf{Boundary Recall} & \textbf{mIoU} & \textbf{P@0.3} & \textbf{P@0.5} & \textbf{P@0.7} \\
\midrule
Uniform & 82.3 & 48.0 & 66.3 & 49.8 & 34.5 \\
\textbf{Event} & \textbf{94.3} & \textbf{50.4} & \textbf{67.6} & \textbf{53.7} & \textbf{35.8} \\
\bottomrule
\end{tabular}
}
\end{minipage}
\hfill
\begin{minipage}[t]{0.49\linewidth}
\centering
\caption{
Comparison of the proposed embedding matching with discrete time token prediction in RTL performance.}
\vspace{-2mm}
\label{tab:embedding_grounding}
\resizebox{\linewidth}{!}{
\begin{tabular}{lccccc}
\toprule
\textbf{Method} & \multicolumn{2}{c}{\textbf{Step 1}} & \multicolumn{3}{c}{\textbf{Step 2}} \\
\cmidrule(lr){2-3} \cmidrule(lr){4-6}
& mIoU & P@0.5 & mIoU & P@0.5 & GPT-4 \\
\midrule
Discrete token prediction
& 45.7 & 45.9 & 40.9 & 40.2 & 36.8 \\
\textbf{Embedding matching} & \textbf{49.2} & \textbf{48.9} & \textbf{50.4} & \textbf{53.7} & \textbf{42.4} \\
\bottomrule
\end{tabular}
}
\end{minipage}

\vspace{-4mm}
\end{table*}

\subsection{Quantitative Analysis}

\noindent
\textbf{ActivityNet-RTL.}
We evaluate EventCoT on ActivityNet-RTL~\citep{lita_ECCV_2024}.
We employ mIoU and P@$K$ as localization accuracy metrics and GPT-4 score to evaluate answer quality, following LITA~\citep{lita_ECCV_2024}.
\arredit{We first evaluate recent video QA models, including large-scale foundation models, in the zero-shot setting, with detailed prompts and evaluation protocol in Appendix~\ref{supple:foundation_baselines}.
However, even these large-scale foundation models still struggle on this task, showing that reasoning temporal localization remains highly challenging and far from solved.
Furthermore, fine-tuning the latest open-source video QA models such as Qwen3.5-9B~\citep{qwen3.5} and TimeLens-8B~\citep{timelens} on our training data improves their temporal localization but degrades their answer quality, indicating that a single VLM struggles to optimize the two objectives jointly.
EventCoT instead separates temporal grounding from answer generation, so that each sub-task is optimized without competing with the other.
As shown in Table~\ref{tab:lita_benchmark}, even with the Vicuna-7B backbone, EventCoT outperforms all compared models on every metric while using less than $20\%$ of the visual tokens of LITA.
Replacing the backbone with Qwen2.5-7B further improves the results, achieving the best performance overall.}

\noindent
\textbf{ReXTime.}
\youngkil{To further assess the generalization of our model, we evaluate its zero-shot performance on ReXTime~\citep{chen2024rextime}.
\arredit{We use R@$K$ and mIoU as localization metrics, and Accuracy (Acc) and Accuracy@IoU$\geq$0.5 (Acc@IoU) for answering, where Acc@IoU requires both a correct answer and a predicted moment with IoU of at least $0.5$, jointly reflecting answering and localization quality.}
As shown in Table~\ref{tab:zs_rextime}, EventCoT achieves the best performance with the Qwen2.5-7B backbone across all metrics.
Even with a weaker backbone such as Vicuna-7B, EventCoT remains the best among models with comparable backbones, and its temporal localization even surpasses prior works built on stronger backbones.
This demonstrates that EventCoT's effectiveness stems from the method rather than the backbone.}

\subsection{In-depth Analysis}

We verify the effectiveness of EventCoT through in-depth analysis on the ActivityNet\hspace{0pt}-RTL~\citep{lita_ECCV_2024} benchmark. 
To enable extensive experiments at scale, all ablations are trained for $10\%$ of the full fine-tuning schedule.

\noindent
\arredit{\textbf{Necessity of Event-centric Tokenization.}
As shown in Table~\ref{tab:uniform_vs_event}, replacing event-centric intervals with uniform intervals reduces boundary recall and degrades temporal localization performance, confirming that aligning intervals with semantic event transitions is essential for reliable localization.}

\noindent
\arredit{\textbf{Necessity of Embedding Matching.}
Table~\ref{tab:embedding_grounding} shows that substituting embedding matching with discrete time-token prediction leads to performance drops in both Step~1 and Step~2, and lower GPT-4 score, indicating weaker temporal reasoning.}

\begin{figure}[!tp]
    \centering
    \includegraphics[width=1.0\columnwidth]{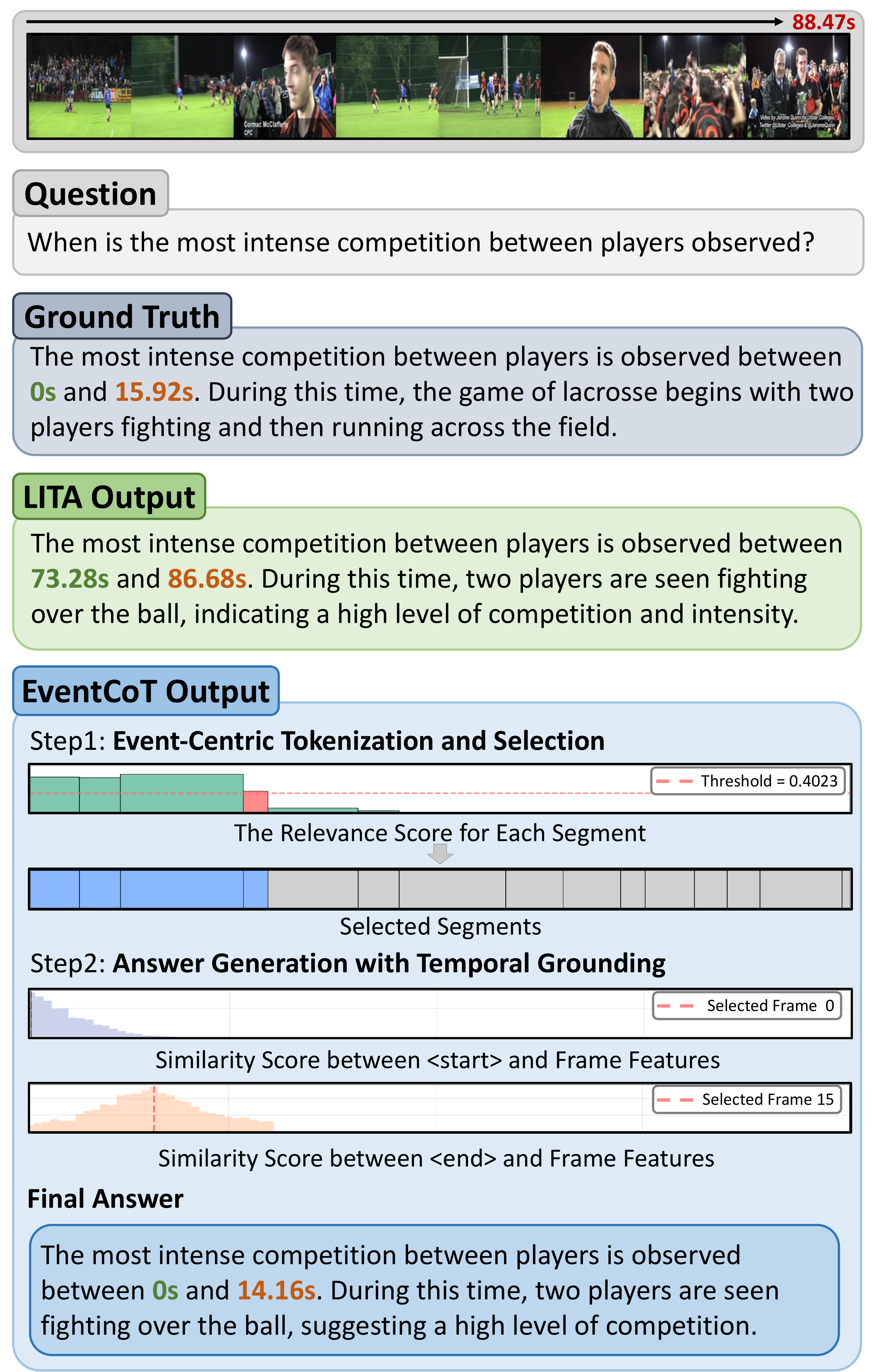}
    \caption{Qualitative results on ActivityNet-RTL~\cite{lita_ECCV_2024}.}
    \label{fig:main_qual}
    \vspace{-0.4cm}
\end{figure}

\begin{figure}[!tp]
    \centering
    \includegraphics[width=1.0\columnwidth]{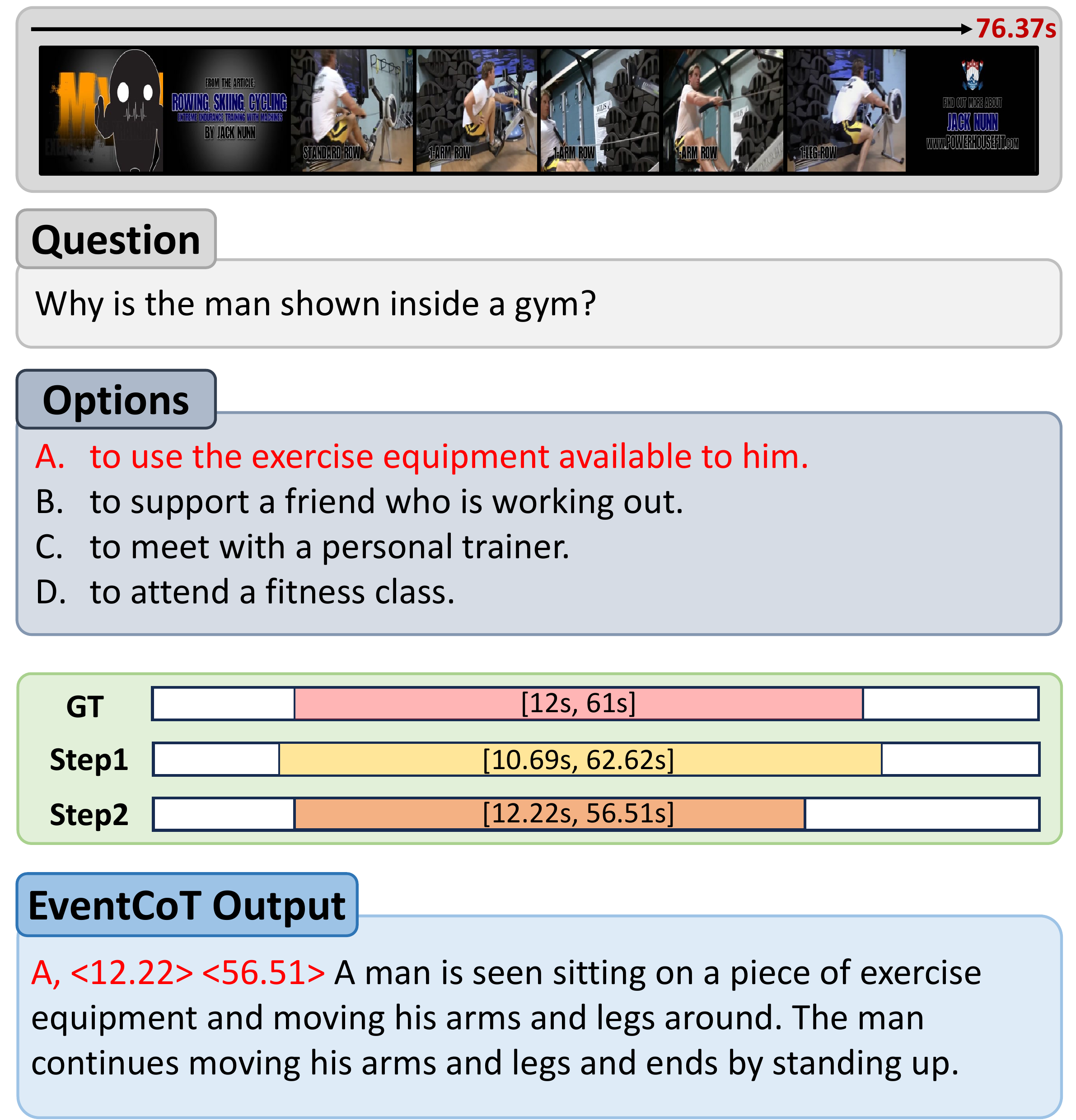}
    \vspace{-0.5cm}
    \caption{Zero-shot qualitative results on ReXTime~\cite{chen2024rextime}.}
    \label{fig:main_qual_2}
    \vspace{-5mm}
\end{figure}
\arredit{Further analyses, including the effect of individual components, the impact of each loss term, additional zero-shot evaluations, and our event boundary detection with full comparisons, hyperparameter ablations, feature discriminability, and inference time, are provided in Appendix~\ref{supple:ablation}.}

\subsection{Qualitative Results}
Fig.~\ref{fig:main_qual} illustrates the outputs of EventCoT and LITA~\citep{lita_ECCV_2024}. 
\arredit{EventCoT first identifies question-relevant events and predicts start and end timestamps via embedding matching.}
As shown in the example, event-centric video CoT reasoning enables EventCoT to produce markedly more accurate temporal intervals and more reliable answers than LITA for reasoning temporal localization.
Fig.~\ref{fig:main_qual_2} further presents zero-shot qualitative results on ReXTime~\citep{chen2024rextime}.
\arredit{In this example, EventCoT localizes query-relevant moments and selects the correct answer candidate, indicating effective temporally grounded reasoning beyond the training distribution.
Additional qualitative results are provided in Appendix~\ref{supple:qual}.}

\section{Conclusion}

We have presented EventCoT, an event-centric, token-efficient chain-of-thought framework for reasoning temporal localization. 
\arredit{By leveraging event-centric tokenization with a \cotmark{chain-of-thought process}, EventCoT effectively identifies question-relevant events and performs fine-grained reasoning.}
\youngkil{Extensive experiments on ActivityNet-RTL for reasoning temporal localization and on ReXTime for grounded video question answering demonstrate substantial improvements in both temporal grounding and answer quality, validating the effectiveness and efficiency of our method.} 
\clearpage
\section*{Limitations}
\youngkil{EventCoT fixes the number of events to $N$ and uniformly samples $T=100$ frames per video, which specializes the framework to relatively short videos.
Adapting the number of events and frames to the length of each video is therefore a natural direction for future work.
In addition, each placeholder token is currently optimized by the language modeling loss and the embedding matching loss at once, so a single embedding has to serve both generation and matching.
Fully separating these roles would require techniques such as latent CoT~\citep{latentcot}, which generates multiple latent embeddings when producing a single next token.}

\appendix

\setcounter{section}{0}
\renewcommand{\thesection}{\Alph{section}}
\setcounter{table}{0}
\renewcommand{\thetable}{S\arabic{table}}
\setcounter{figure}{0}
\renewcommand{\thefigure}{S\arabic{figure}}

\clearpage

This supplementary material provides additional details and analyses of EventCoT.
\arredit{The contents are organized as follows.}

\startcontents[appendix]
\printcontents[appendix]{l}{1}{\setcounter{tocdepth}{2}}
\section{Training and Inference Pipeline} 
\label{supple:process_detail}

This section provides additional details on how EventCoT is trained jointly in a multi-turn manner and how its \cotmark{two-step process} is executed during inference.

\subsection{Training Procedure}

During training, EventCoT performs question-relevant event selection (Step~1) and fine-grained reasoning with relevant events (Step~2) within a single multi-turn conversation. 
Because ground-truth temporal intervals are available during training, both stages can be supervised jointly.

\noindent\textbf{Question-relevant Event Selection (Step~1).}
The model receives the event tokens and the input question, and is prompted to output the special token \ptok{segment\_key}. 
The hidden embedding at this token position is supervised using the event selection loss $\mathcal{L}_\text{sel}$, computed from the overlap with the ground-truth interval.

\noindent\textbf{Fine-grained Reasoning with Relevant Events (Step~2).}
In the next turn, the model receives the frame-level tokens, constructed from both the selected events, determined by overlap with the ground-truth interval during training, and a randomly sampled subset of unselected events, and generates an answer that includes \ptok{start} and \ptok{end} placeholder tokens. 
The hidden embeddings at these token positions are trained using the temporal localization loss $\mathcal{L}_\text{ts}$, based on the ground-truth start and end timestamps.

\subsection{Inference Procedure}

Unlike training, inference is executed sequentially: Step~1 first identifies the question-relevant events, and only then Step~2 performs fine-grained temporal reasoning based on those selected events. 

\noindent\textbf{Question-relevant Event Selection (Step~1).}
The model receives the event tokens and the input question, produces \ptok{segment\_key}, and computes its embedding similarity against the event tokens to determine the question-relevant events.

\noindent\textbf{Fine-grained Reasoning with Relevant Events (Step~2).}
\youngkil{Step~2 proceeds as the next turn of the multi-turn conversation.
Its input comprises the frame-level tokens of the selected events, augmented as described below, the input question, and the preserved Step~1 turn, which still carries both the event tokens of the whole video and the \ptok{segment\_key} output.
Through this multi-turn context, Step~2 inherits the event-level representation of the entire video, so regions outside the selected interval, such as causes and consequences, remain accessible as global context. Meanwhile, the frame-level tokens let the model zoom into the selected events for fine-grained reasoning.}
\youngkil{At inference, Step~1 selections are no longer guaranteed to match the ground-truth events.
To keep Step~2's input consistent between training and inference, we apply the same sampling scheme used in training, augmenting the selected events with a randomly sampled subset of unselected events (Eq.~\eqref{eq:visual_tokens}).
During training, the random sampling exposes Step~2 to events beyond the ground-truth selection. Applying the same scheme at inference bridges the gap with Step~1's noisy predictions.}

\youngkil{EventCoT then generates the answer together with the \ptok{start} and \ptok{end} placeholder tokens, reusing the reasoning signal carried over from Step~1.
At these positions, it performs embedding matching~\citep{liu2024bench,lai2024lisa} between the placeholder token embeddings and those of the frame-level visual tokens to infer the start and end timestamps.}

\section{Additional Method Details}
\label{supple:method_details}

\arredit{\subsection{Distance Factor}}
\arredit{The original distance factor $\delta$~\citep{dpc-knn} quantifies the relative isolation of each frame from those with higher $\rho$:
\begin{equation}
\label{eq:distance_factor}
\delta_i =
\begin{cases}
\min_{j : \rho_j > \rho_i} d(i, j), & \text{if } \exists j: \rho_j > \rho_i, \\
\max_{j} d(i, j), & \text{otherwise.}
\end{cases}
\end{equation}
Non-peak frames typically have a nearby denser neighbor (small $\delta$), while only local density maxima stay far from any denser frame (large $\delta$), yielding fewer and more meaningful peaks per cluster.}

\arredit{The normalization in Eq.~\eqref{eq:power_scaling} is defined as $\mathrm{Normalize}(\delta_i) = \max\left(0, (\delta_i - P_{10})/(P_{90} - P_{10})\right)$, where $P_{10}$ and $P_{90}$ denote the 10th and 90th percentiles of the distance factors within the video.}

\arredit{\subsection{Extended Attention Scope}}
\arredit{The extended index set in Eq.~\eqref{eq:event_feature} is defined as $\mathcal{E}_n^{\text{ext}} = \{i \mid i \in [\max(1, \mathrm{idx}({\mathbf{e}_n})[0] - \beta), \min(T, \mathrm{idx}({\mathbf{e}_n})[-1] + \beta)]\}$, with $\mathrm{idx}({\mathbf{e}_n})$ denoting the sorted frame indices of the $n$-th event.}

\arredit{\subsection{DIoU Loss}}
\arredit{The DIoU loss between the predicted interval $[\hat{t}^{\text{start}}, \hat{t}^{\text{end}}]$ and the ground-truth interval is defined as
\begin{equation}
\label{eq:diou_loss}
\mathcal{L}_{\text{diou}} = 1 - \text{IoU}(\hat{t}, t_{\text{gt}}) + \frac{d^2}{c^2},
\end{equation}
where $d$ and $c$ denote the distance between interval centers and the smallest enclosing 1-D box length.}

\section{Additional Experimental Details}
\label{supple:exp_details}

\arredit{\subsection{Training Datasets}}
\arredit{EventCoT is trained on a mixture of video-language datasets with and without temporal annotations.
Datasets with temporal annotations include those for dense video captioning as well as that for RTL.
In the case of the RTL dataset, ActivityNet-RTL~\citep{lita_ECCV_2024}, we use its training split with 33,557 question--answer pairs.
We also adopt the training splits of the video captioning datasets ActivityNet-Captions~\citep{krishna2017dense} and YouCook2~\citep{zhou2018towards}; in particular, for learning RTL using these datasets, we formulate an event localization task by treating each caption as a question and predicting its corresponding time interval.
Datasets without time intervals include those for video question answering and image-based instruction tuning.
Specifically, we use NExT-QA~\citep{xiao2021next}, a video question answering dataset with 37K videos, and LLaVA-150K~\citep{liu2023visual}, an instruction tuning dataset containing 150K multi-turn instructions.
To utilize these datasets for training, we train Step~1 to predict the \ptok{segment\_key} and train Step~2 to generate answers using all video frame features.
Since no ground-truth time intervals are available, training on these datasets is done solely with the language modeling loss.}

\arredit{\subsection{Hyperparameters}}
\arredit{For the Vicuna-7B backbone, we set $k=5$ for event boundary detection using DPC-KNN.
The event context attention module consists of a single attention layer with $n_{\text{heads}}=32$, hidden dimension $4096$, and temporal extension factor $\beta=3$, which is trained from scratch.
We use a temperature of $\tau=0.07$, set the selection threshold to $\gamma=0.5$, set the sampling ratio of $\mathcal{R}$ to $30\%$, and set the loss coefficients to $\lambda_{\text{sel}}=0.1$ and $\lambda_{\text{ts}}=0.1$.
For the Qwen2.5-7B backbone, we initialize from VideoLLaMA3-7B~\citep{videollama3}, follow its hidden dimension of $3584$ with $28$ attention heads in the event context attention, and set the sampling ratio of $\mathcal{R}$ to $20\%$ and the selection threshold to $\gamma=0.4$, with all other hyperparameters unchanged.}

\arredit{\subsection{Training Details}}
\arredit{EventCoT is trained for 500K iterations with replacement using a learning rate of $\expnum{2}{5}$ and a warm-up ratio of $0.03$.
Training is conducted on 4 RTX~6000~ADA GPUs, taking about 30~hours with a batch size of~4 for the Vicuna-7B backbone.
The Qwen2.5-7B variant is trained on the same GPUs with LoRA (rank 16), a learning rate of $\expnum{2}{4}$, and a batch size of~2, taking about 41~hours.}

\arredit{\subsection{Benchmark Details}}
\arredit{ActivityNet-RTL is derived from the validation split of ActivityNet-Captions~\citep{krishna2017dense}, where GPT-4~\citep{chatgpt} generates reasoning-based ``when'' questions from temporally localized dense captions, followed by manual filtering and verification.
ReXTime is constructed from the ActivityNet~\citep{krishna2017dense} and QVHighlights~\citep{lei2021detecting} datasets, which provide videos with time-aligned captions describing temporally localized events.
GPT-4 models~\citep{chatgpt,2023GPT4VisionSC} generate reasoning-based multiple-choice question--answer pairs from temporally related event pairs, and the answer temporal spans are subsequently re-annotated and verified by human annotators.
We evaluate ReXTime in a zero-shot manner.
Neither its training split nor its multiple-choice format is used during training, and its evaluation videos do not overlap with our training videos.}

\arredit{\subsection{Licenses of External Assets}}
\begin{table*}[t]
\centering
\small
\caption{Licenses and sources for all external assets used in this paper.}
\label{tab:licenses}
\arredit{%
\resizebox{\textwidth}{!}{%
\begin{tabular}{l l l p{6.4cm}}
\toprule
\textbf{Asset} & \textbf{Type} & \textbf{License} & \textbf{Source} \\
\midrule
ActivityNet-RTL~\citep{lita_ECCV_2024} & Dataset & Nvidia Source Code License-NC & \url{https://github.com/NVlabs/LITA} \\
ActivityNet-Captions~\citep{krishna2017dense} & Dataset & Unspecified (research use) & \url{https://cs.stanford.edu/people/ranjaykrishna/densevid/} \\
YouCook2~\citep{zhou2018towards} & Dataset & Unspecified (research use) & \url{http://youcook2.eecs.umich.edu} \\
NExT-QA~\citep{xiao2021next} & Dataset & Unspecified (research use) & \url{https://github.com/doc-doc/NExT-QA} \\
LLaVA-Instruct-150K~\citep{liu2023visual} & Dataset & CC BY 4.0 & \url{https://huggingface.co/datasets/liuhaotian/LLaVA-Instruct-150K} \\
ReXTime~\citep{chen2024rextime} & Dataset & CC BY-NC-SA 4.0 & \url{https://github.com/ReXTime/ReXTime} \\
\midrule
CLIP ViT-L/14~\citep{clip_pmlr21} & Model & MIT & \url{https://github.com/openai/CLIP} \\
Vicuna-7B~\citep{vicuna2023} & Model & Llama 2 Community License & \url{https://huggingface.co/lmsys/vicuna-7b-v1.5} \\
LLaVA~\citep{liu2023visual} & Model & Apache 2.0 & \url{https://github.com/haotian-liu/LLaVA} \\
LITA~\citep{lita_ECCV_2024} & Model & CC BY-NC-SA 4.0 (weights) & \url{https://github.com/NVlabs/LITA} \\
Qwen2.5-7B & Model & Apache 2.0 & \url{https://huggingface.co/Qwen} \\
Qwen3.5-9B~\citep{qwen3.5} & Model & Apache 2.0 & \url{https://huggingface.co/Qwen} \\
TimeLens-8B~\citep{timelens} & Model & Custom (academic use only) & \url{https://github.com/TencentARC/TimeLens} \\
GPT-4~\citep{chatgpt} & Model (API) & OpenAI Terms of Use & \url{https://openai.com/policies/} \\
\midrule
PySceneDetect~\citep{scenedetect2025} & Software & BSD-3-Clause & \url{https://scenedetect.com} \\
\bottomrule
\end{tabular}%
}}
\end{table*}

\arredit{Table~\ref{tab:licenses} summarizes the licenses and sources of all external datasets, models, and software used in this work.}

\section{Additional In-depth Analysis}
\label{supple:ablation}

\begin{table*}[!t]
\centering

\begin{minipage}[t]{0.49\linewidth}
\centering
\caption{Effect of individual components.}
\vspace{-2mm}
\label{tab:module_ablation}
\resizebox{\linewidth}{!}{
\begin{tabular}{lccccc}
\toprule
\textbf{Method}  & \textbf{mIoU} & \textbf{P@0.3} & \textbf{P@0.5} & \textbf{P@0.7} & \textbf{GPT-4} \\
\midrule
\textbf{Ours}   & \textbf{50.4} & \textbf{67.6} & \textbf{53.7} & \textbf{35.8} & 42.4 \\
w/o event context attention        & 46.4 & 62.4 & 48.5 & 31.9 & 34.1 \\
w/o sampled events $\mathcal{R}$ & 36.1 & 49.3 & 33.6 & 10.9 & 39.0 \\
w/ sampled frames                  & 40.8 & 57.2 & 40.6 & 21.0 & \textbf{42.8} \\
\bottomrule
\end{tabular}
}
\end{minipage}
\hfill
\begin{minipage}[t]{0.49\linewidth}
\centering
\caption{Impact of the loss functions.}
\vspace{-2mm}
\label{tab:loss_ablation}
\resizebox{\linewidth}{!}{
\begin{tabular}{lcccccccc}
\toprule
\multirow{2}{*}{\textbf{Method}} 
& \multicolumn{5}{c}{\textbf{Step 1}} &&
\multicolumn{2}{c}{\textbf{Step 2}} \\ 
\cmidrule(lr){2-6} \cmidrule(lr){8-9}
& \textbf{Precision} & \textbf{Recall} & \textbf{Accuracy} 
& \textbf{mIoU} & \textbf{P@0.5} &&
\textbf{mIoU} & \textbf{P@0.5} \\
\midrule
\textbf{Ours} & \textbf{66.0} & 91.7 & \textbf{83.2} & \textbf{49.2} & \textbf{48.9} && \textbf{50.4} & \textbf{53.7} \\
w/o $\mathcal{L}_\text{sel}$   & 30.2 & \textbf{100.0} & 30.2 & 26.0 & 16.6 && 26.1 & 16.6 \\
w/o $\mathcal{L}_\text{ce}$    & 45.1 & 89.9 & 63.9 & 32.6 & 28.4 && 18.1 & 13.5 \\
w/o $\mathcal{L}_\text{diou}$  & 62.4 & 91.7 & 80.7 & 46.0 & 43.7 && 45.7 & 48.0 \\
\bottomrule
\end{tabular}
}
\end{minipage}

\vspace{-4mm}
\end{table*}

\arredit{\subsection{Effect of Individual Components}}
\arredit{Table~\ref{tab:module_ablation} presents the contribution of each proposed component.
Removing the event context attention and using event point features instead of event tokens degrades both localization and answer generation quality. This indicates that event representations enriched beyond simple mean-pooled features are crucial for reliable event selection and downstream reasoning.
Omitting the randomly sampled events $\mathcal{R}$ from unselected ones leads to significant performance drops especially in temporal grounding, since it is overly tied to event boundaries seen during training and fails to generalize.
Finally, replacing these sampled events $\mathcal{R}$ with sampled frames slightly improves GPT-4 score, presumably because frame-level sampling offers broader visual context.
However, this comes at the cost of significantly worse temporal grounding, as frame-level sampling introduces substantial noise to focus on temporally coherent regions.}

\arredit{\subsection{Impact of the Loss Function}}
\arredit{Table~\ref{tab:loss_ablation} presents the effect of each loss component on temporal grounding.
In Step~1, we report event-level metrics (precision, recall, and accuracy) to evaluate the quality of question-relevant event selection.
Both $\mathcal{L}_\text{sel}$ in Step~1 and $\mathcal{L}_\text{ce}$ in Step~2 are essential; removing either leads to a drastic performance drop across all metrics.
Moreover, removing $\mathcal{L}_\text{diou}$ degrades performance, indicating that modeling the dependency between start and end timestamps is beneficial.}

\arredit{\subsection{Zero-shot Temporal Grounding on Charades-STA}}
\label{supple:charades}
\begin{table}[t]
\centering
\small
\caption{Zero-shot temporal grounding on the Charades-STA~\citep{gao2017tall} test split. FT denotes fine-tuning on Charades-STA. $^\dagger$Zero-shot results reported by \citet{timelens}.}
\label{tab:charades_zeroshot}
\arredit{%
\resizebox{\columnwidth}{!}{%
\begin{tabular}{lccccc}
\toprule
\textbf{Method} & \textbf{FT} & \textbf{R1@0.3} & \textbf{R1@0.5} & \textbf{R1@0.7} & \textbf{mIoU} \\
\midrule
VideoChat-R1-7B~\citep{videochatr1} & \cmark & - & 71.7 & 50.2 & 60.8 \\
TempSamp-R1-7B~\citep{tempsampr1} & \cmark & 83.6 & 74.1 & 52.9 & 62.1 \\
\midrule
GPT-4o~\citep{gpt4o}$^\dagger$ & \xmark & 51.6 & 27.9 & 11.7 & 34.7 \\
Qwen2.5-VL-7B~\citep{qwen2.5-vl}$^\dagger$ & \xmark & 59.4 & 39.8 & 14.5 & 43.6 \\
\midrule
EventCoT (Vicuna-7B) & \xmark & 85.7 & 65.3 & 34.4 & 57.3 \\
EventCoT (Qwen2.5-7B) & \xmark & \textbf{96.6} & \textbf{87.4} & \textbf{59.0} & \textbf{70.9} \\
\bottomrule
\end{tabular}%
}}
\end{table}

\arredit{To further examine the generality of the two-step design, we evaluate EventCoT on the Charades-STA~\citep{gao2017tall} test split without any training on Charades.
Each annotation is converted into our event localization format, where the query sentence is rephrased as a question asking when the described activity happens, and the model predicts the start and end timestamps with the same two-step pipeline and hyperparameters used for RTL.
As shown in Table~\ref{tab:charades_zeroshot}, both backbones surpass the zero-shot baselines by large margins, and the Qwen2.5-7B variant exceeds even the methods fine-tuned on Charades-STA across all metrics, indicating that event-centric two-step grounding transfers beyond its training domains.}

\arredit{\subsection{Zero-shot Moment Retrieval on QVHighlights}}
\label{supple:qvh}
\begin{table}[t]
\centering
\small
\caption{Zero-shot moment retrieval on the QVHighlights~\citep{lei2021detecting} validation split. FT denotes fine-tuning on QVHighlights. Bold marks the best result among methods not fine-tuned on QVHighlights.}
\label{tab:qvh_zeroshot}
\arredit{%
\resizebox{0.9\columnwidth}{!}{%
\begin{tabular}{lccc}
\toprule
\textbf{Method} & \textbf{FT} & \textbf{R1@0.5} & \textbf{R1@0.7} \\
\midrule
Moment-DETR~\citep{lei2021detecting} & \cmark & 59.7 & 40.8 \\
QD-DETR~\citep{qddetr} & \cmark & 62.7 & 46.7 \\
CG-DETR~\citep{cgdetr} & \cmark & 67.4 & 52.1 \\
Chrono-BLIP~\citep{chrono} & \cmark & 74.8 & 60.5 \\
LLaVA-MR~\citep{llavamr} & \cmark & 78.1 & 64.1 \\
\midrule
Moment-GPT~\citep{momentgpt} & \xmark & 58.9 & 38.6 \\
Chrono-GPT~\citep{chrono} & \xmark & 61.7 & 41.8 \\
GranAlign~\citep{granalign} & \xmark & 61.9 & 41.8 \\
\midrule
EventCoT (Vicuna-7B) & \xmark & 48.2 & 28.6 \\
EventCoT (Qwen2.5-7B) & \xmark & \textbf{71.7} & \textbf{47.4} \\
\bottomrule
\end{tabular}%
}}
\end{table}

\arredit{We further evaluate EventCoT on the QVHighlights~\citep{lei2021detecting} validation split under the same zero-shot transfer protocol as Charades-STA, and score the predictions with the official evaluation code, which measures Recall@1 against all annotated moments of each query.
As shown in Table~\ref{tab:qvh_zeroshot}, the Qwen2.5-7B variant surpasses the best zero-shot baseline and the fine-tuned task-specific models, and narrows the gap to the LLM-based models fine-tuned on QVHighlights.
The Vicuna-7B variant transfers less effectively through its Step~2 timestamps on this benchmark.
In contrast, the interval from its Step~1 event selection reaches an R1@0.5 of $63.5$ and an R1@0.7 of $41.7$, again exceeding the best zero-shot baseline, suggesting that event selection generalizes more robustly than timestamp generation under domain shift.}

\arredit{\subsection{Full Comparison of Event Boundary Detection Methods}}

\begin{table}[t]
\centering
\caption{Full comparison of event boundary detection methods.}
\label{tab:boundary_detection_methods}
\begin{subtable}[t]{\linewidth}
\centering
\vspace{-2mm}
\arredit{%
\resizebox{\columnwidth}{!}{%
\begin{tabular}{llc}
\toprule
\textbf{Method}  & \textbf{Local Boundary Score} & \textbf{Avg. Recall} \\
\midrule
\multirow{5}{*}{\textbf{KNN}~\citep{1053964}}
        & Gaussian density                  & 64.6 \\
        & Euclidean mean                    & 66.6 \\
        & Local Euclidean mean              & 55.5 \\
        & Local Euclidean std.              & 57.0 \\
        & Euclidean mean \& Local Euclidean std. & 58.1 \\
\midrule
\multirow{5}{*}{\textbf{DPC-KNN}~\citep{dpc-knn}}
        & Gaussian density                  & 86.2 \\
        & Euclidean mean                    & 89.9 \\
        & Local Euclidean mean              & 89.3 \\
        & Local Euclidean std.              & 84.5 \\
        & Euclidean mean \& Local Euclidean std. & 89.1 \\
\midrule
\multirow{4}{*}{\textbf{Ours}}
        & Euclidean mean                          & 89.9 \\
        & Euclidean mean (+ power scaling)        & 93.9 \\
        & Euclidean mean (+ temporal proximity)   & 92.6 \\
        & Euclidean mean (+ both)                 & \textbf{94.3} \\
\bottomrule
\end{tabular}%
}}
\caption{Local boundary score variants.}
\label{tab:boundary_detection_variants_full}
\end{subtable}
\vspace{1mm}
\begin{subtable}[t]{\linewidth}
\centering
\arredit{%
\resizebox{\columnwidth}{!}{%
\begin{tabular}{lccc}
\toprule
\textbf{Method} & \textbf{Avg. Recall} & \textbf{Mem. (MB)} & \textbf{T (ms)} \\
\midrule
\textbf{Ours} & \textbf{94.3} & \textbf{0.78} & \textbf{0.44} \\
Strefer~\citep{zhou2025strefer} & 88.0 & 29.49 & 86.53 \\
PySceneDetect~\citep{scenedetect2025} & 62.5 & 28.71 & 76.53 \\
\bottomrule
\end{tabular}%
}}
\caption{Comparison with prior methods.}
\label{tab:boundary_detection_prior}
\end{subtable}
\end{table}

\arredit{Table~\ref{tab:boundary_detection_variants_full} reports the full comparison of local boundary score variants under KNN and DPC-KNN.
Boundary recall measures the fraction of ground-truth boundary points, namely the start and end timestamps of each ground-truth interval, that are matched by the detected boundaries.
A point counts as matched when it falls within a window of 5\% of the video duration centered on a detected boundary, and the recall is aggregated over all evaluation samples.
KNN-based methods, which rely solely on local boundary scores, show limited performance regardless of whether Euclidean or other local metrics are used.
Adopting DPC-KNN improves the average recall up to $89.9$, confirming that incorporating the distance factor is crucial for identifying event boundaries.
Among different local boundary scores, Euclidean mean shows the best result under both KNN and DPC-KNN, and serves as the basis for our design.
Building on this, our two key improvements, \textit{temporal proximity-aware distance} and \textit{power-scaled distance factor}, consistently enhance the performance, and combining both achieves the best average recall of $94.3$, demonstrating their complementary effect.
Furthermore, as shown in Table~\ref{tab:boundary_detection_prior}, our method outperforms pixel-change-based boundary detection approaches such as PySceneDetect~\citep{scenedetect2025} and Strefer~\citep{zhou2025strefer}.
This is because our detection method targets semantic event transitions rather than pixel-level shot changes, enabling accurate boundary localization even without explicit cuts, while remaining substantially more memory- and time-efficient.}

\subsection{Feature Discriminability Analysis}

\begin{table*}[t]
\centering
\caption{
Feature discriminability in step~1 (mean~$\pm$~std). 
Event feature similarity measures intra-sample similarity between event features within the same video, 
while segment key similarity measures inter-sample similarity across different videos. 
Lower similarity indicates higher discriminability. The model is trained for 10\% of the full fine-tuning schedule.
}
\label{sub_tab:feature_discriminability}
\resizebox{\textwidth}{!}
{
\begin{tabular}{lcc}
\toprule
\textbf{Feature Type} & \textbf{Inter-event Similarity} & \textbf{GT vs. Non-GT Similarity} \\
\midrule
Event point feature (before event context attention)& 
0.6228 {\tiny$\pm$ 0.0278} & 
0.5766 {\tiny$\pm$ 0.0504} \\

Event token (after event context attention)& 
0.5832 {\tiny$\pm$ 0.0332} & 
0.5337 {\tiny$\pm$ 0.0551} \\

Event embedding (after LLM)& 
0.4481 {\tiny$\pm$ 0.0330} & 
0.3364 {\tiny$\pm$ 0.1034} \\

Event embedding (after projector) & 
0.7961 {\tiny$\pm$ 0.0182} & 
0.7353 {\tiny$\pm$ 0.0635} \\
\midrule
\textbf{Segment Key Discriminability}& \multicolumn{2}{c}{\textbf{Inter-sample Similarity}} \\
\midrule
Segment key embedding (after LLM) & 
\multicolumn{2}{c}{0.6870 {\tiny$\pm$ 0.0947}} \\
Segment key embedding (after projector) & 
\multicolumn{2}{c}{0.9155 {\tiny$\pm$ 0.0624}} \\
\bottomrule
\end{tabular}}
\end{table*}
While our embedding matching effectively handles event selection and temporal grounding (Table~\ref{tab:lita_benchmark}), we further examine how discriminative the features remain throughout the overall pipeline.
We also analyze the impact of applying projectors,
each implemented as a separate 2-layer MLP applied to the event embeddings and the placeholder token embeddings (\ptok{segment\_key}, \ptok{start}, \ptok{end}) immediately before embedding matching.
We empirically find that removing these projectors leads to better performance in EventCoT.

\begin{table}[t]
\centering
\caption{Ablation study on the projection head in step~1 temporal localization. All models are trained for 10\% of the full fine-tuning schedule.}
\label{sub_tab:projection_head_comparison}
{
\resizebox{\columnwidth}{!}{%
\begin{tabular}{lcccc}
\toprule
\textbf{Method} & \textbf{mIoU} & \textbf{P@0.3} & \textbf{P@0.5} & \textbf{P@0.7} \\
\midrule
w/o Projector &
\textbf{49.6} & \textbf{63.3} & \textbf{48.9} & \textbf{34.9} \\

w/ Projector &
33.3 & 48.0 & 30.1 & 13.5 \\
\bottomrule
\end{tabular}}
}
\end{table}

Table~\ref{sub_tab:feature_discriminability} reports the cosine similarities among event features across four stages: \textit{event point feature} (before event context attention), \textit{event token} (after event context attention), \textit{event embedding} (after LLM), and \textit{projected event embedding} (after the projector).
We observe a consistent trend in which event features become increasingly more discriminative as they progress from event point features to event tokens and finally to event embeddings. This is reflected in decreasing overall similarity and reduced similarity between question-relevant (GT) and non-relevant events.
These results indicate that the model gradually refines event representations in a way that better supports embedding matching for question-relevant event selection.

However, applying the projectors markedly increases feature similarity across all metrics, indicating a collapse in discriminability.
This trend also appears in the segment-key discriminability analysis in Table~\ref{sub_tab:feature_discriminability}, where inter-sample similarity rises sharply after passing through the projector.
This issue becomes especially problematic under our Step~1 temporal localization setup, where time intervals are constructed from consecutive question-relevant events, and the longest consecutive interval is selected as the final output.
Under this evaluation protocol, training with projectors leads to a substantial performance drop (Table~\ref{sub_tab:projection_head_comparison}), demonstrating that projectors degrade the model’s ability to distinguish event features.
These results justify removing the projectors to preserve feature discriminability and improve the reliability of embedding matching.

\begin{table}[t]
\centering
\caption{Ablation study on the number of events for temporal localization. All models are trained for 10\% of the full fine-tuning schedule.}
\label{tab:num_segments_ablation}
{
\resizebox{\columnwidth}{!}{%
\begin{tabular}{lcccc}
\toprule
\textbf{\# Events (N)} & \textbf{mIoU} & \textbf{P@0.3} & \textbf{P@0.5} & \textbf{P@0.7} \\
\midrule
8  & 39.5 & 55.4 & 38.5 & 17.9 \\
16 & 50.4 & 67.6 & 53.7 & 35.8 \\
32 & \textbf{53.0} & \textbf{72.9} & \textbf{56.8} & \textbf{39.7} \\
64 & 47.5 & 66.3 & 53.7 & 30.6 \\
100 & 47.6 & 72.0 & 54.2 & 21.4 \\
\bottomrule
\end{tabular}}}
\end{table}

\subsection{Effect of the Number of Events}
\label{sup_sec:num_seg}

We conduct an ablation study on the number of events $N$. 
Using fewer events reduces the computational cost of event selection but risks merging multiple semantic events into a single segment, making it harder to identify question-relevant regions. 
Conversely, increasing the number of events prevents multiple events from being merged, but may fragment a single coherent event across different segments and introduce redundant computation.

As shown in Table~\ref{tab:num_segments_ablation}, \youngkil{the choice of $N$ affects temporal localization more than the other hyperparameters, since $N$ determines the semantic granularity of events.
Both $N{=}16$ and $N{=}32$ achieve competitive results, indicating that moderate values of $N$ balance accuracy and efficiency.}
We adopt $N{=}16$ in our experiments as it provides strong performance while maintaining lower computational cost in Step~1. 
Since both too few and too many events degrade performance, these results highlight the importance of determining the number of events adaptively rather than fixing it for all videos.
Developing an adaptive event segmentation strategy is therefore a promising future direction for improving both accuracy and efficiency.

\begin{table}[t]
\centering
\caption{Ablation study on the $\beta$ of event context attention for temporal localization. All models are trained for 10\% of the full fine-tuning schedule.}
\label{tab:beta_ablation}
{
\begin{tabular}{lcccc}
\toprule
\textbf{Method} & \textbf{mIoU} & \textbf{P@0.3} & \textbf{P@0.5} & \textbf{P@0.7} \\
\midrule
$\beta = 0$     & 50.1 & 67.6 & 51.6 & 33.6 \\
$\beta = 1$    & \textbf{50.8} & \textbf{68.5} & 53.3 & \textbf{36.2} \\
$\beta = 3$    & 50.4 & 67.6 & \textbf{53.7} & 35.8 \\
$\beta = 5$    & 49.2 & 66.3 & 50.7 & 34.5 \\
No mask        & 48.0 & 64.1 & 49.8 & 31.0 \\
\bottomrule
\end{tabular}}
\end{table}
\begin{table}[t]
\centering
\caption{Ablation study on the number of video frames for temporal localization. All models are trained for 10\% of the full fine-tuning schedule.}
\label{tab:num_frames_ablation}
\resizebox{\columnwidth}{!}{%
\begin{tabular}{lcccc}
\toprule
\textbf{\# Video Frames (T)} & \textbf{mIoU} & \textbf{P@0.3} & \textbf{P@0.5} & \textbf{P@0.7} \\
\midrule
50  & 47.5 & 67.2 & 48.1 & 32.7 \\
100 & 50.4 & 67.6 & \textbf{53.7} & \textbf{35.8} \\
200 & 49.8 & 69.8 & 52.4 & 34.9 \\
300 & \textbf{51.4} & \textbf{72.0} & 52.9 & 34.0 \\
\bottomrule
\end{tabular}}
\end{table}

\begin{table*}[t]                                                                   
\centering                                                                         
\caption{Step~2 Performance Analysis by IoU Range}
\label{sub_tab:iou_range_analysis}                                                     
{
\begin{tabular}{lcccccc}                                                           
\toprule                                                                           
IoU Range & Ratio & Step1 mIoU & Step2 mIoU & Step2 P@0.5 & GPT Score \\
\midrule                                                                           
Low (0--33\%) & 26.3\% & 12.9 & 16.8 & 5.0 & 40.2 \\                      
Medium (33--67\%) & 25.4\% & 50.6 & 61.6 & 81.0 & 47.0 \\                 
High (67--100\%) & \textbf{48.2\%} & \textbf{82.6} & \textbf{78.8} & \textbf{92.7} & \textbf{53.7} \\                     
\midrule                                                                           
\textbf{Overall} & 100.0\% & 56.2 & 58.5 & 66.8 & 48.7 \\           
\bottomrule                                                                        
\end{tabular}}
\end{table*}

\subsection{Effect of the Temporal Extension Factor}
To mitigate potential errors in predicted event boundaries, our event context attention allows each event token to attend not only to frames within its own interval but also to a small temporal neighborhood around it. 
We investigate the impact of the temporal extension factor $\beta$, which controls the number of additional frames included on both sides of an event.
Table~\ref{tab:beta_ablation} summarizes the results.
``No mask'' refers to computing attention over all frames, i.e., using the full frame sequence as keys and values without any restriction. 
This causes the model to attend to many irrelevant frames, introducing substantial noise and resulting in worse temporal localization. 
In contrast, masking attention to a local temporal window around each segment yields better localization accuracy. 
Furthermore, including a small number of neighboring frames (e.g., $\beta = 1$ or $3$) consistently improves the performance over using no additional context ($\beta = 0$), confirming that a small amount of boundary-aware context is also beneficial. 
The best performance is obtained at $\beta = 1$, suggesting that our event boundary detector is sufficiently accurate (94.3\% boundary recall), and thus a modest temporal extension is adequate to compensate for false boundary detection without introducing excessive noise.
\youngkil{We adopt $\beta = 3$ in our main experiments, which achieves the best P@$0.5$ and performs comparably to $\beta = 1$ on the other metrics.}

\subsection{Effect of the Number of Video Frames}
We further analyze the effect of the number of sampled video frames $T$.
As shown in Table~\ref{tab:num_frames_ablation}, the performance remains robust across different values of $T$, and even a moderate number of frames (e.g., $T=100$) is sufficient for stable temporal localization. 
This robustness arises from the design of EventCoT: during event-centric tokenization, each event is summarized into a single event embedding, and thus the semantic content of an event is sufficiently captured even with a limited number of frames. 
As long as each event contains enough visual cues, the resulting event embeddings do not change significantly.

Nevertheless, increasing $T$ continues to provide marginal improvements, with the best results obtained at $T=300$. 
This is likely because denser frame sampling offers more reliable local density estimates for event boundary detection, and it also benefits fine-grained timestamp grounding in Step~2, where the similarity between the placeholder tokens and frame features is computed at the frame level. 
These results indicate that EventCoT is inherently robust to the number of sampled frames, yet can still take advantage of additional frame information in longer or more complex videos.

\begin{table*}[t]
\centering
\caption{
Ablation study on the segment selection threshold $\gamma$ for step~1 and step~2 temporal localization.}
\label{tab:step1_threshold_ablation}
\resizebox{\textwidth}{!}
{
\setlength{\tabcolsep}{10pt}
\begin{tabular}{lccccccccc}
\hline
\multirow{2}{*}{\textbf{$\gamma$}} 
& \multicolumn{4}{c}{\textbf{Step 1}} &
& \multicolumn{4}{c}{\textbf{Step 2}} \\ 
\cline{2-5} \cline{7-10}
& \textbf{mIoU} & \textbf{P@0.3} & \textbf{P@0.5} & \textbf{P@0.7} & 
& \textbf{mIoU} & \textbf{P@0.3} & \textbf{P@0.5} & \textbf{P@0.7} \\
\midrule
$0.1$  & 52.0 & 72.9 & 53.7 & 37.6 && 55.5 & 73.8 & 60.2 & 41.0 \\
$0.2$  & 54.0 & 74.7 & 55.9 & 38.9 && 56.7 & 74.2 & 61.1 & 42.8 \\
$0.3$  & 55.6 & 77.3 & 57.6 & 40.6 && 57.5 & 76.0 & 62.9 & 44.1 \\
$0.4$  & 56.4 & 78.6 & 59.0 & 42.4 && 57.6 & 76.9 & 64.2 & 43.7 \\
$\textbf{0.5}$  & 56.2 & 76.9 & 62.0 & 45.4 && \textbf{58.5} & 78.6 & 66.8 & \textbf{45.9} \\
$0.6$  & 57.0 & \textbf{79.5} & 62.9 & 46.3 && 58.2 & \textbf{79.9} & \textbf{67.2} & 44.5 \\
$0.7$  & 57.8 & 79.0 & 63.3 & 49.3 && 56.7 & 78.2 & 65.1 & 43.2 \\
$0.8$  & \textbf{58.7} & \textbf{79.5} & \textbf{65.9} & \textbf{50.2} && 55.3 & 76.9 & 64.6 & 41.5 \\
$0.9$  & 57.9 & 77.3 & 65.1 & 48.9 && 52.3 & 74.2 & 59.0 & 38.0 \\
\bottomrule
\end{tabular}
}
\end{table*}

\subsection{Effect of Step~1 Localization on Step~2 Accuracy}

To examine how event selection quality in Step~1 affects the final temporal grounding performance, we divide the validation samples into three groups based on the IoU between the predicted intervals of Step~1 and the ground-truth. 
The predicted interval is computed as the union of consecutive selected events; if multiple candidates exist, we select the one containing the largest number of events. 
Table~\ref{sub_tab:iou_range_analysis} summarizes the results.
Nearly half of the samples (48.2\%) fall into the \textit{High} IoU range ($>$67\%), showing that Step~1 reliably identifies question-relevant events. 
As expected, higher Step~1 accuracy leads to better Step~2 temporal grounding, since precise event selection restricts the candidate frames for embedding matching.

The \textit{High} IoU group shows a slight decrease in Step~2 mIoU. 
Since our detected event boundaries are highly accurate (94.3\% recall), Step~1 often yields intervals already close to the ground-truth. 
In such cases, because the additional Step~2 refinement is performed jointly with answer generation, it may introduce errors in the predicted intervals and marginally lower IoU.
For the \textit{Low} IoU group ($<$33\%), Step~1 fails to localize well, yet Step~2 still finds some temporal boundaries (5.0\% P@$0.5$) and increases mIoU. 
This comes from two factors: (1) incorporating randomly sampled unselected events, which provides global context, and (2) frame-level embedding matching, which can identify relevant cues even outside the mis-localized events.

Overall, these results demonstrate that Step~1 and Step~2 form a complementary coarse-to-fine process. 
Accurate Step~1 predictions improve fine-grained reasoning, while additional global context allows Step~2 to partially recover when Step~1 is inaccurate.

\subsection{Effect of the Event Selection Threshold}
We ablate the threshold $\gamma$ used for selecting question-relevant events in Step~1. 
To ensure that at least one event is passed to Step~2, the threshold is defined as $\gamma \cdot \max_n r_n$, where $r_n$ denotes the predicted relevance score. 
As shown in Table~\ref{tab:step1_threshold_ablation}, temporal localization performance remains stable across a wide range of $\gamma$ values (0.1–0.9). 
This robustness indicates that the model assigns sharply higher relevance scores to true relevant events, making the separation between relevant and non-relevant events robust to threshold variation. 
Based on this observation, we simply adopt $\gamma = 0.5$ for main experiments.

\begin{table}[t]
\centering
\caption{Ablation on the sampling ratio of additional unselected events provided to the LLM in Step~2.}
\label{tab:threshold_ablation}
\begin{tabular}{lcccc}
\toprule
\textbf{Ratio} & \textbf{mIoU} & \textbf{P@0.3} & \textbf{P@0.5} & \textbf{P@0.7} \\
\midrule
w/o $\mathcal{R}$ & 36.1 & 49.3 & 33.6 & 10.9 \\
10\% & 50.9 & 69.9 & 52.4 & 32.3 \\
20\% & \textbf{52.4} & \textbf{72.5} & \textbf{53.7} & 35.4 \\
30\% & 50.4 & 67.6 & \textbf{53.7} & \textbf{35.8} \\
40\% & 47.4 & 65.1 & 48.0 & 32.3 \\
50\% & 40.3 & 57.2 & 41.5 & 22.3 \\
\bottomrule
\end{tabular}
\end{table}

\begin{figure*}[t]
    \centering
    \includegraphics[width=1\textwidth]{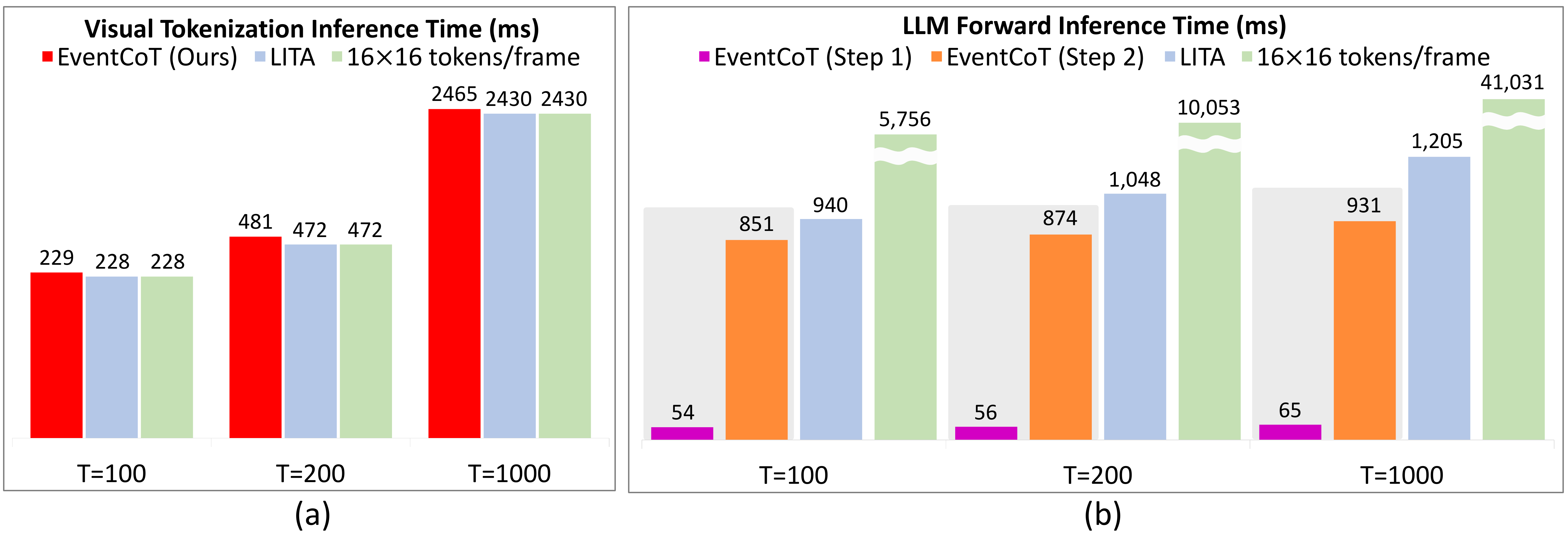}
    \caption{Inference time analysis. (a) Visual tokenization time for different video lengths. (b) LLM forward inference time.}
    \label{fig:supple_inference_time}
\end{figure*}
\subsection{Effect of the Sampling Ratio}
We conduct an ablation on the sampling ratio of additional unselected events $\mathcal{R}$ that are randomly incorporated into the Step~2 input.
As shown in Table~\ref{tab:threshold_ablation}, including $\mathcal{R}$ during Step~2 training substantially improves temporal localization compared to removing $\mathcal{R}$.
This is because $\mathcal{R}$ breaks the boundary alignment between the Step~2 input intervals and the ground-truth interval, mitigating a shortcut where the model simply outputs the input interval boundaries as the predicted start and end timestamps rather than performing genuine temporal reasoning.
However, overly large ratios (e.g., $\geq 40\%$) degrade performance, as excessive unselected events dilute the question-relevant evidence identified by Step~1 and introduce additional noise, making fine-grained temporal grounding less reliable.
\arredit{Based on this observation, we adopt a sampling ratio of $30\%$ for the Vicuna-7B backbone and $20\%$ for the Qwen2.5-7B backbone.}

\subsection{Effect of Temporal Proximity Weight $\alpha$}
\begin{table}[t]
\centering
\caption{Boundary recall on ActivityNet-RTL under varying temporal proximity weight $\alpha$ in event boundary detection. The recall stays within a narrow band across the entire sweep, showing that $\alpha$ is robust rather than hard-coded; we use $\alpha = 0.3$.}
\label{tab:alpha_sweep}
\setlength{\tabcolsep}{6pt}
\resizebox{\columnwidth}{!}{%
\begin{tabular}{lcccccc}
\toprule
$\alpha$ & 0.0 & 0.1 & 0.2 & 0.3 (ours) & 0.4 & 0.5 \\
\midrule
Boundary Recall (\%) & 93.9 & 94.1 & 94.1 & 94.3 & 93.9 & \textbf{94.5} \\
\bottomrule
\end{tabular}}
\end{table}
\youngkil{We ablate the temporal proximity weight $\alpha$, which weights temporal proximity against feature distance in the pairwise frame distance used for event boundary detection.
As shown in Table~\ref{tab:alpha_sweep}, the boundary recall stays within a narrow band of $0.6$ percentage points across the entire sweep $\alpha \in [0.0, 0.5]$.
Even its lowest value still exceeds \youngkil{the recalls of} competing boundary detection methods such as Strefer~\citep{zhou2025strefer} and PySceneDetect~\citep{scenedetect2025}\youngkil{, which are reported in Table~\ref{tab:boundary_detection_prior}}.
The recall is highest at $\alpha = 0.5$, yet we adopt $\alpha = 0.3$ without any tuning, since the method is insensitive to this hyperparameter rather than relying on a carefully chosen value.}

\subsection{Robustness to Hyperparameters}
\youngkil{Most hyperparameters of EventCoT exhibit wide robust ranges.
Across the temporal extension factor $\beta$, the number of frames $T$, the event selection threshold $\gamma$, the sampling ratio of additional unselected events, and the temporal proximity weight $\alpha$, performance stays within a narrow margin over a wide range of values.
The main exception is the number of events $N$, which determines the semantic granularity of events; too few events merge distinct semantics into one segment while too many fragment a coherent event, so $N$ should be chosen at a moderate scale, where both $N{=}16$ and $N{=}32$ perform competitively.
\arredit{We therefore use a single default configuration per backbone in all experiments without tuning any hyperparameter per dataset.}}

\subsection{In-depth Inference Time Analysis}
Figure~\ref{fig:supple_inference_time} presents a detailed analysis of the inference time across different steps of the pipeline.
As shown in Fig.~\ref{fig:supple_inference_time}(a), the visual tokenization time is comparable across all three methods.
In the case of EventCoT, a slight increase in computation is observed because the event tokenizer additionally performs event context attention when constructing event tokens.
Nevertheless, the overall visual tokenization time remains similar to that of the baseline methods.
In contrast, Fig.~\ref{fig:supple_inference_time}(b) shows that EventCoT achieves substantially faster LLM forward inference compared to the other approaches.
Because each event interval is summarized into a compact event token, Step~1 requires only a small number of tokens, which significantly reduces the inference time.
Furthermore, Step~1 effectively identifies question-relevant events, allowing Step~2 to operate only on a subset of the video rather than the entire sequence.
As a result, the number of visual tokens processed by the LLM is greatly reduced, leading to more efficient inference while preserving the information necessary for accurate answer generation and temporal grounding.

\section{Additional Qualitative Results}
\label{supple:qual}

\begin{figure*}[t]
    \centering
    \includegraphics[width=\textwidth]{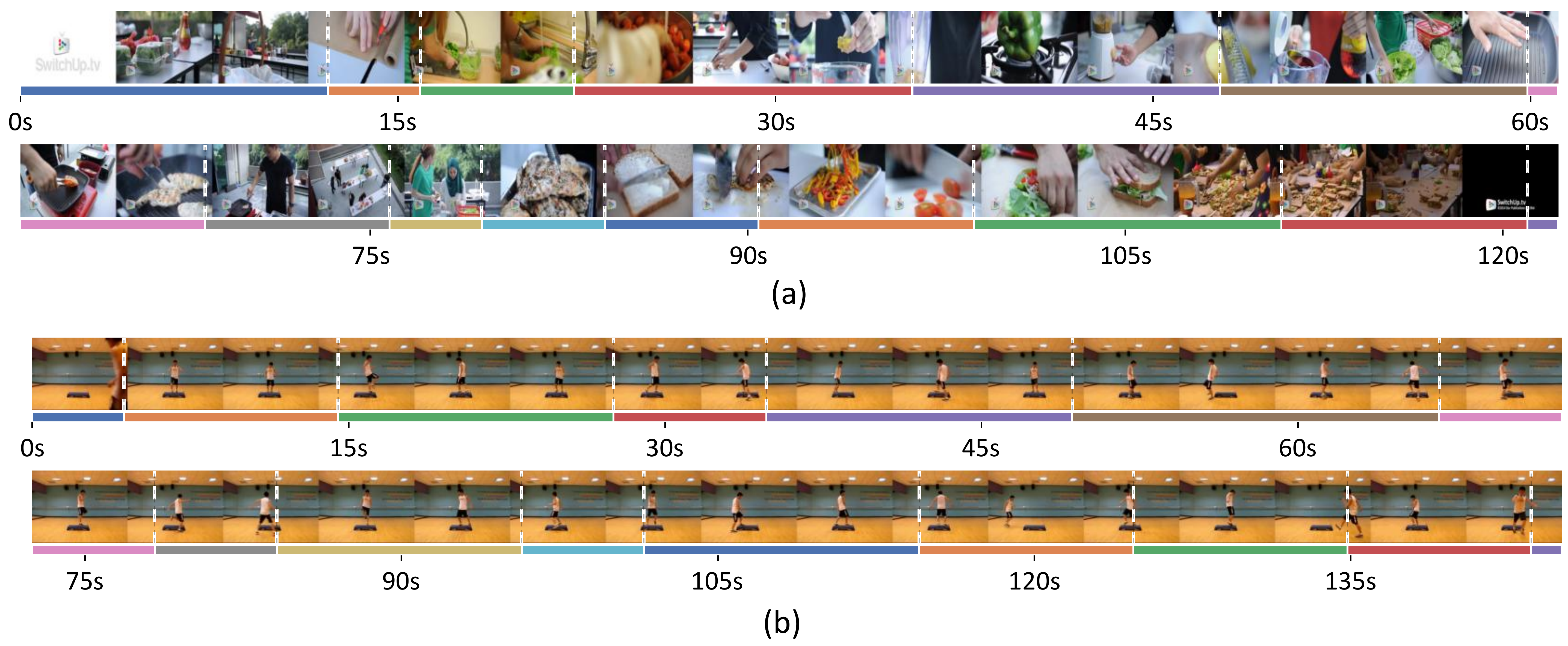}
    \caption{\youngkil{Qualitative examples of event segmentation on ActivityNet-RTL.
    Uniformly sampled frames are shown along the time axis with the detected event boundaries (white dashed lines) and the resulting events (colored bars).
    (a) A cooking video with frequent scene changes.
    (b) A single-shot exercise video without explicit scene cuts.}}
    \label{fig:supple_seg_qual}
\end{figure*}

\noindent\youngkil{\textbf{Event segmentation quality.}
Figure~\ref{fig:supple_seg_qual} visualizes the event boundaries detected by our event tokenizer on two ActivityNet-RTL videos.
In the first video, which contains frequent scene changes, the detected boundaries closely follow the transitions between cooking steps such as ingredient preparation, blending, and grilling, so each event forms a semantically coherent unit.
The second video is captured by a single fixed camera without any shot change, yet the boundaries still partition the video into distinct phases of the exercise, since our boundary detection relies on semantic feature distances rather than pixel-level changes.
These examples show that the event tokenizer yields event units aligned with semantic transitions in both multi-shot and single-shot videos, providing a reliable basis for event-level selection in Step~1.}

\begin{figure*}[t]
    \centering
    \includegraphics[width=0.95\textwidth]{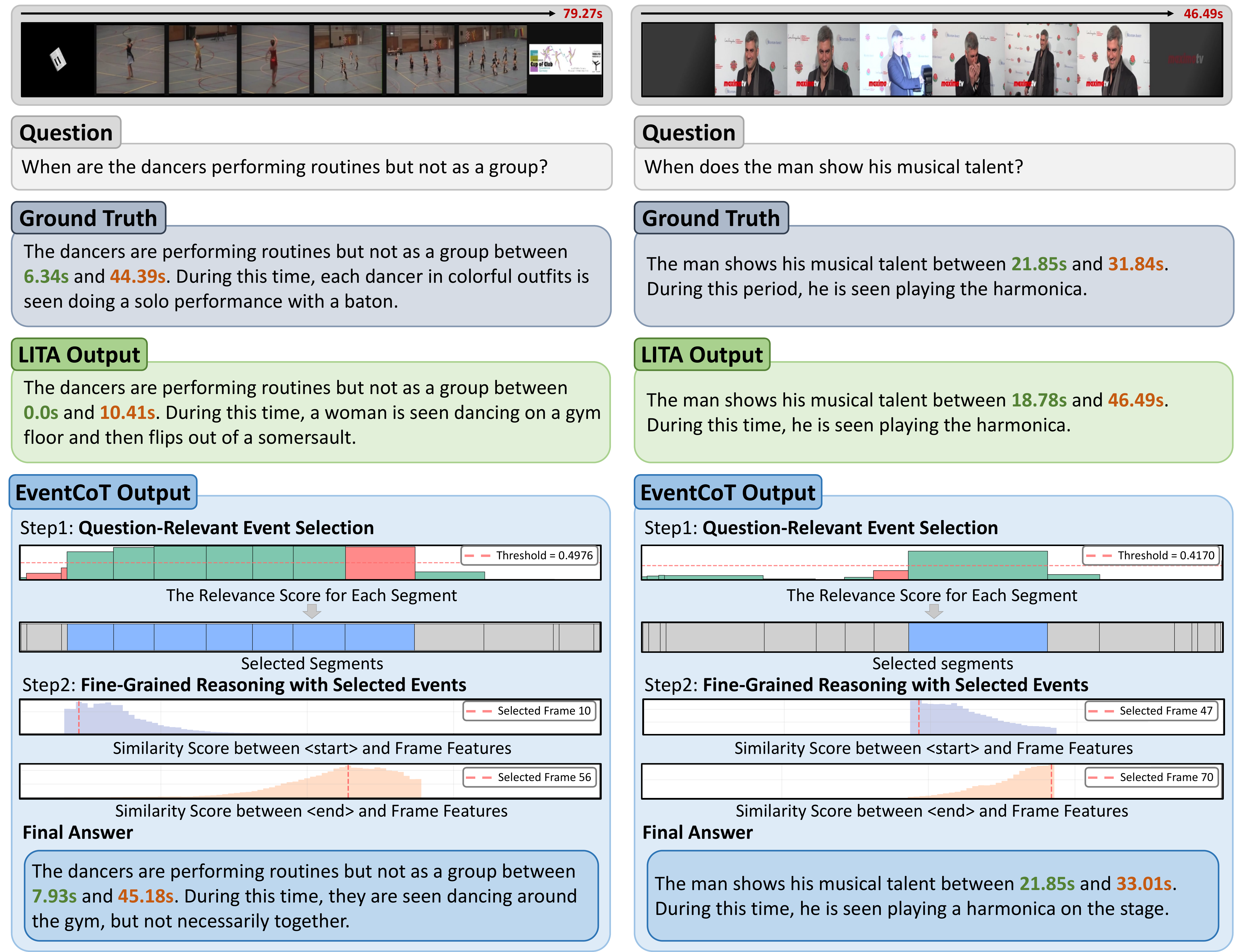}
    \caption{Additional Qualitative Results of LITA~\cite{lita_ECCV_2024} and EventCoT on the ActivityNet-RTL~\cite{lita_ECCV_2024}.}
    \label{fig:supple_qual_1}
\end{figure*}

\begin{figure*}[t]
    \centering
    \includegraphics[width=0.95\textwidth]{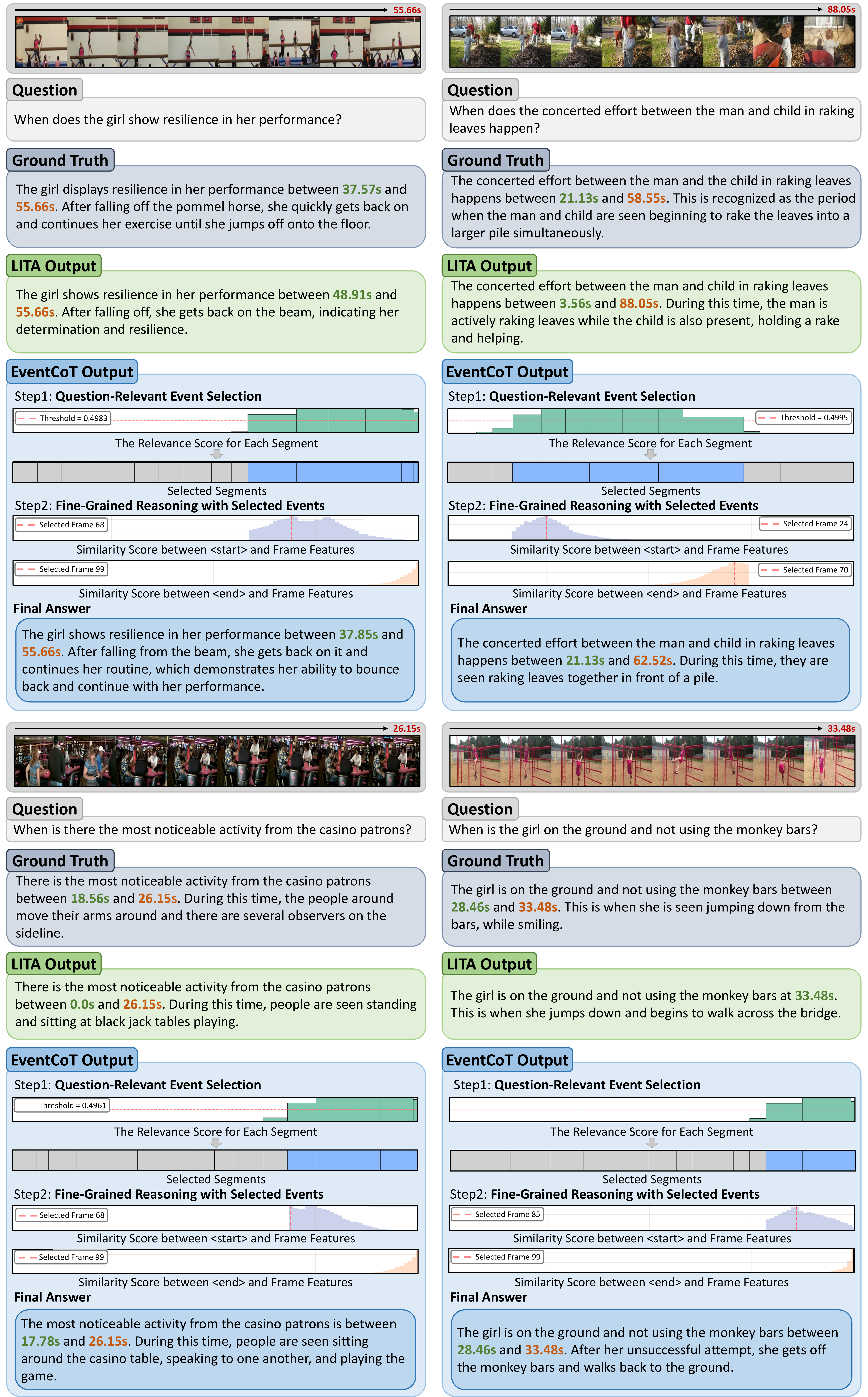}
    \caption{\textbf{Additional Qualitative Results of LITA~\cite{lita_ECCV_2024}} and \textbf{EventCoT on the ActivityNet-RTL~\cite{lita_ECCV_2024}.}}
    \label{fig:supple_qual_2}
\end{figure*}

\begin{figure*}[t]
    \centering
    \includegraphics[width=0.95\textwidth]{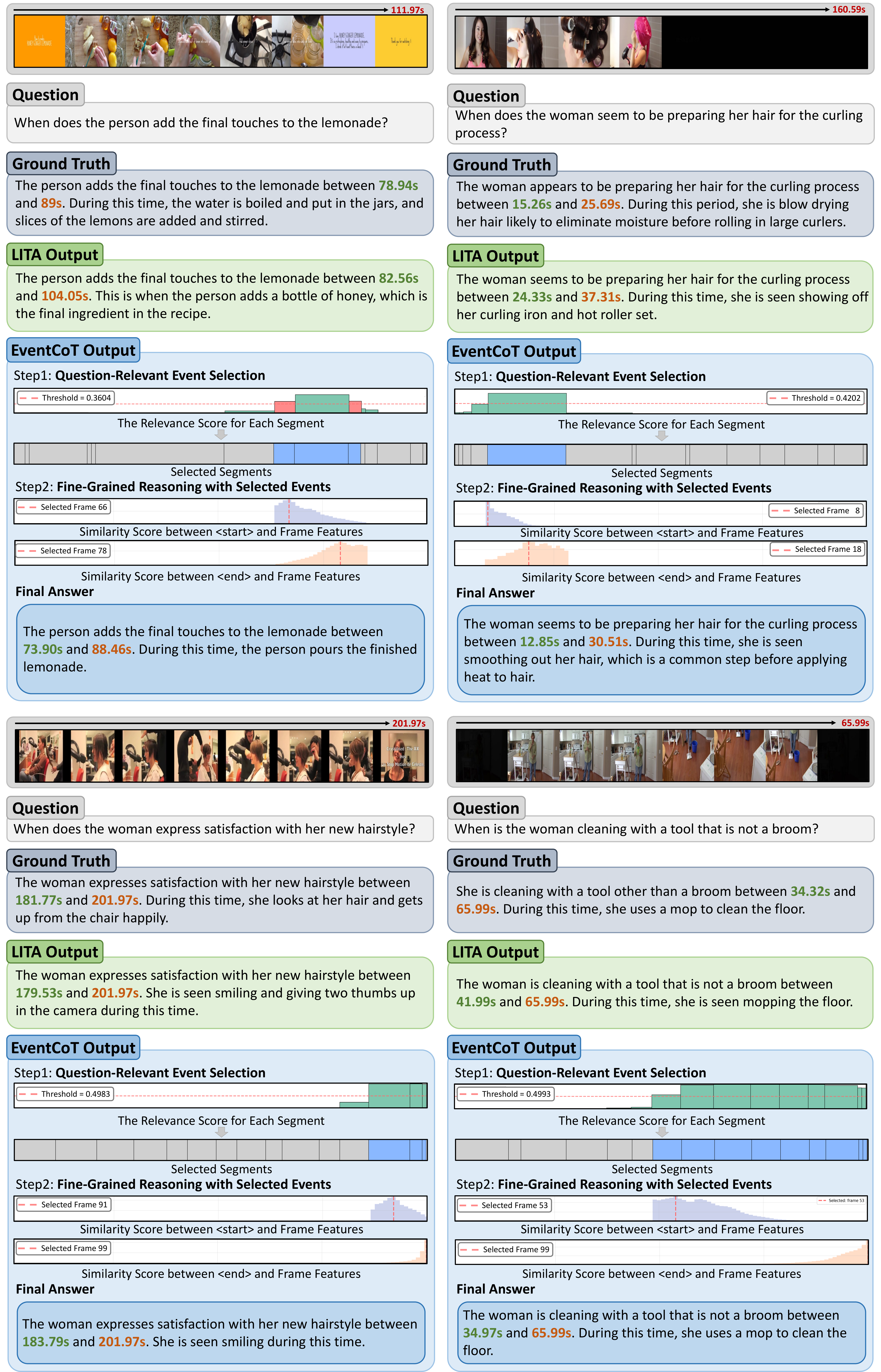}
    \caption{\textbf{Additional Qualitative Results of LITA~\cite{lita_ECCV_2024}} and \textbf{EventCoT on the ActivityNet-RTL~\cite{lita_ECCV_2024}.}}
    \label{fig:supple_qual_3}
\end{figure*}

\noindent Figures~\ref{fig:supple_qual_1}–\ref{fig:supple_qual_3} provide additional qualitative comparisons between LITA~\citep{lita_ECCV_2024} and EventCoT on ActivityNet-RTL.
Across these examples, we observe that EventCoT performs robustly across a wide range of temporal positions. 
Because both stages of our framework rely on embedding matching grounded in the semantic information contained in the visual features, EventCoT accurately localizes events occurring at early, middle, or late positions in the video. 
Moreover, it consistently generates appropriate and contextually relevant answers to the given questions, demonstrating strong temporal reasoning and semantic understanding.

\section{Task Prompts}
\label{supple:task_prompt}
This section presents the task-specific prompts used in Step~1 and Step~2.
\youngkil{All tasks follow the same two-step prompt template shown below: Step~1 uses the \emph{Event Selection Prompt}, and Step~2 uses the \emph{Fine-grained Reasoning Prompt}.}
From this shared template, we instantiate the template fields \texttt{\{task\_prompt\}}, \texttt{\{question\_prompt\}}, \texttt{\{step1\_prompt\}}, and \texttt{\{step2\_prompt\}} with task-specific instructions and question texts, producing the final prompts used for both training and inference.
\youngkil{In the prompts, the L-th hidden state refers to the final-layer hidden embedding of the LLM.}

\noindent\youngkil{\textbf{Event Selection Prompt (Step~1):}}
\begin{lstlisting}[style=prompt]
Task: {task_prompt}
Event features: <image>
Question: {question_prompt}
Instruction: {step1_prompt}
\end{lstlisting}

\noindent\youngkil{\textbf{Fine-grained Reasoning Prompt (Step~2):}}
\begin{lstlisting}[style=prompt]
Task: {task_prompt}
Selected frame features: <image>
Question: {question_prompt}
Instruction: {step2_prompt}
\end{lstlisting}

\subsection{Dense Video Captioning}

The prompt construction for dense video captioning~\citep{zhou2018towards,krishna2017dense} follows LITA~\citep{lita_ECCV_2024}, where each question is formed by combining a task description with a timestamp instruction. 
The task description specifies the overall task objective, while the timestamp instruction indicates how temporal boundaries should be expressed. 
\youngkil{A question is generated by randomly sampling one option from each pool of task descriptions and timestamp instructions and concatenating them.
The listing below shows the instantiated \texttt{\{task\_prompt\}}, \texttt{\{question\_prompt\}}, \texttt{\{step1\_prompt\}}, and \texttt{\{step2\_prompt\}} for dense video captioning.}

\begin{lstlisting}[style=prompt]
task_prompt:
- Dense Video Captioning

Task Descriptions (sample one):
- Provide a detailed description of the given video.
- Describe the provided video in detail.
- Summarize the visual content of the video.
- Write an informative summary of the video.

Timestamp Instructions (sample one):
- Each sentence should begin with the start and end timestamps.
- At the beginning of each sentence, include the start and end timestamps.
- Prepend each sentence with its start and end timestamps.

question_prompt:
- (sampled Task Description) + " " + (sampled Timestamp Instruction)

step1_prompt:
- Based on the given event features and question, predict a <segment_key> to select question-relevant event segments. The <segment_key> will be matched with the L-th hidden state embedding of event features through embedding matching to identify relevant event segments.

step2_prompt:
- Using the given event features and selected frame features, provide a detailed answer to the question.
\end{lstlisting}

\subsection{Event Localization}

Event localization~\citep{zhou2018towards,krishna2017dense} aims to predict the temporal interval in which a given event description occurs within the video. 
To construct questions for this task, we follow the same strategy as dense video captioning: a task description is combined with a timestamp instruction. 
Here, the task description is formed by inserting the given sentence into a template that queries when the event happens in the video, while the timestamp instruction specifies the format of the desired temporal boundaries.
\youngkil{The instantiated \texttt{\{task\_prompt\}}, \texttt{\{question\_prompt\}}, \texttt{\{step1\_prompt\}}, and \texttt{\{step2\_prompt\}} for event localization are shown below.}

\begin{lstlisting}[style=prompt]
task_prompt:
- Event Localization

Task Descriptions (sample one):
- When does "{SENTENCE}" happen in the video?
- At what point in the video does "{SENTENCE}" happen?
- When is "{SENTENCE}" depicted in the video?
- At what time in the video does "{SENTENCE}" take place?

Timestamp Instructions (sample one):
- Provide the start and end timestamps when the event occurs.
- Answer with the start and end timestamps of the event.
- Indicate when the event happens using start and end timestamps.

question_prompt:
- (sampled Task Description) + " " + (sampled Timestamp Instruction)

step1_prompt:
- Based on the given event features and question, predict a <segment_key> to select question-relevant event segments. The <segment_key> will be matched with the L-th hidden state embedding of event features through embedding matching to identify relevant event segments.

step2_prompt:
- Using the given event features and selected frame features, provide the start and end timestamps when the event occurs.
\end{lstlisting}

\subsection{Other Tasks: RTL, Video-QA, and Image-based Instruction Tuning}

For the remaining tasks, namely reasoning temporal localization (RTL~\citep{lita_ECCV_2024}), video question answering (Video-QA~\citep{xiao2021next}), and image-based instruction tuning~\citep{liu2023visual}, the question is directly provided by the dataset without additional construction. 
All three tasks share the same Step~1 prompt, which predicts the \ptok{segment\_key} using the event features and the input question.
In Step~2, the prompt is slightly adapted for each task to reflect its output format. 
For RTL, the Step~2 instruction asks the model to produce an answer accompanied by temporal boundaries. 
For Video-QA, the model is prompted to generate a standard answer to the question. 
For image-based instruction tuning, the instruction requests \textit{detailed answers}, since multiple QA pairs may be generated for a single image.
\youngkil{We therefore show each task's \texttt{\{task\_prompt\}} and task-specific \texttt{\{step2\_prompt\}} below.}

\noindent\youngkil{\textbf{(a) Reasoning Temporal Localization}}
\begin{lstlisting}[style=prompt]
task_prompt:
- Reasoning Temporal Localization

step2_prompt:
- Using the given event features and selected frame features, provide a detailed answer to the question with timestamps.
\end{lstlisting}

\noindent\youngkil{\textbf{(b) Video Question Answering}}
\begin{lstlisting}[style=prompt]
task_prompt:
- Video Question Answering

step2_prompt:
- Using the given event features and selected frame features, provide an answer to the question.
\end{lstlisting}

\noindent\youngkil{\textbf{(c) Image-based Instruction Tuning}, whose deployed \texttt{\{task\_prompt\}} is phrased as visual question answering.}
\begin{lstlisting}[style=prompt]
task_prompt:
- Visual Question Answering

step2_prompt:
- Using the given event features and selected frame features, provide detailed answers to the question.
\end{lstlisting}

\subsection{Prompts and Evaluation Protocol for Foundation-Model Baselines}
\label{supple:foundation_baselines}
\youngkil{We evaluate every zero-shot baseline under a single protocol that follows the ActivityNet-RTL task.
Each model is given the video, its total duration, and the question, and is asked to return one line containing both a free-form answer and the start and end timestamps of the supporting interval.
Stating the duration lets the model report timestamps in absolute seconds within the actual length of the video.
From each response we extract the two timestamps, clamp them to the valid range, and use the timestamps to compute mIoU and P@$K$, and the answer to compute the GPT-4 score.
All models share the prompt template below, where the duration and question are filled in for each sample.}

\begin{lstlisting}[style=prompt]
Video duration: {duration} seconds.
Question: {question}

Output exactly ONE line in this format:
The event happens between <START> and <END> seconds. <DESCRIPTION>

Where <START> and <END> are floating-point seconds (0 to {duration}), and <DESCRIPTION> is one sentence describing what happens during that interval to answer the question.
\end{lstlisting}

\youngkil{The models differ in how the video is supplied.
Gemini~2.5 Pro and Gemini~2.5 Flash receive the full video through the Gemini File API, where the video is decoded at roughly one frame per second together with its audio track.
The GPT models (GPT-5, GPT-5-mini, and GPT-4o), Qwen3.5-9B, and TimeLens-8B instead receive $100$ frames sampled uniformly across the video, matching the frame budget of EventCoT.
The GPT models encode each frame in the low-detail setting, which rescales it to $512\times512$ and represents it with a fixed budget of $85$ tokens, so all frames use the same number of visual tokens.
Qwen3.5-9B receives the frames at their native resolution, whereas TimeLens-8B applies a total-pixel budget over the frame sequence that caps its visual token count.}

\section{Future Work}
\label{supple:future_work}
While EventCoT substantially improves temporal localization and reasoning for RTL, several promising directions remain for further advancement.

\noindent
\textbf{(1) Handling variable-length videos.}
\youngkil{We uniformly sample 100 frames per video following LITA~\citep{lita_ECCV_2024}.
Letting EventCoT adjust the number of processed frames to the video duration, or even operate in a streaming setting, would better align it with real-world long videos.}

\noindent\textbf{(2) Adaptive event boundary detection.}
\youngkil{As discussed in Sec.~\ref{sup_sec:num_seg}, the optimal number of events varies across videos, while we fix it to $N$.
Since our detector already produces reliable boundary scores, these scores could be used to infer the number of boundaries per video, potentially improving temporal localization.}

\noindent\textbf{(3) Latent CoT for more fine-grained reasoning.}
\youngkil{The placeholder tokens \ptok{segment\_key}, \ptok{start}, and \ptok{end} are optimized by the language modeling loss and their matching losses at once, which limits the specialization of each token.
Incorporating \emph{latent CoT}~\citep{latentcot}, where a placeholder triggers a short latent reasoning trace dedicated to constructing a more discriminative matching embedding, could decouple these roles and improve both localization and reasoning.}

\noindent\textbf{(4) Incorporating spatially detailed visual features.}
\youngkil{EventCoT relies on spatially aggregated features, which limits queries that require fine-grained spatial understanding.
Integrating spatial feature maps, for instance within the fine-grained reasoning of Step~2, could extend EventCoT to a broader range of vision-language tasks.}

\noindent\textbf{(5) Hierarchical event selection.}
\youngkil{EventCoT selects question-relevant events once before fine-grained reasoning.
A hierarchical process that progressively narrows the relevant temporal regions could further improve localization and reduce noise from irrelevant content.}

\end{document}